\documentclass[journal]{IEEEtran/IEEEtran}
\ifCLASSINFOpdf
   \usepackage[pdftex]{graphicx}
\else
\fi
\usepackage{colortbl}
\usepackage[usenames,dvipsnames]{xcolor}
\usepackage{color}
\usepackage{algorithm}
\usepackage{multirow}
\usepackage{microtype}
\usepackage[bookmarks=false]{hyperref}

\usepackage[noend]{algpseudocode}
\algnewcommand\algorithmicforeach{\textbf{for each}}
\algdef{S}[FOR]{ForEach}[1]{\algorithmicforeach\  #1\ \algorithmicdo}

\newcolumntype{L}[1]{>{\arraybackslash}m{#1}}
\newcolumntype{C}[1]{>{\centering\arraybackslash}m{#1}}

\usepackage{subfig}
\usepackage{amsthm}
\definecolor{myblue}{rgb}{0.65,0.75,0.85}
\definecolor{myblue_light}{rgb}{0.85,0.9,1}
\usepackage{tcolorbox}
\tcbuselibrary{theorems}
\newtcbtheorem[number within=section]{definition}{\small{Definition}}%
{colback=myblue_light,colframe=myblue,fonttitle=\bfseries,coltitle=black}{th}

\begin{document}
%
\title{Quality and Diversity Optimization:\\ A Unifying  Modular Framework}
%
%
%

\author{Antoine~Cully~and~Yiannis~Demiris,~\IEEEmembership{Senior~Member,~IEEE}
\thanks{A. Cully and Y. Demiris are with the Personal
    Robotics Laboratory, Department of Electrical and Electronic
    Engineering, Imperial College London, U.K.
    (e-mail: a.cully@imperial.ac.uk; y.demiris@imperial.ac.uk).}}

\maketitle

\begin{abstract}
  The optimization of functions to find the best solution
  according to one or several objectives has a central role
  in many engineering and research fields. Recently, a new family of
  optimization algorithms, named Quality-Diversity optimization, has
  been introduced, and contrasts with classic algorithms. Instead of
  searching for a single solution, Quality-Diversity algorithms are
  searching for a large collection of both diverse and high-performing
  solutions. The role of this collection is to cover the range of possible solution types as much as
  possible, and to contain the
  best solution for each type. The contribution of this paper is
  threefold. Firstly, we present a unifying framework of
  Quality-Diversity optimization algorithms that covers the two main
  algorithms of this family (Multi-dimensional Archive of Phenotypic
  Elites and the Novelty Search with Local Competition), and that
  highlights the large variety of variants that can be investigated within
  this family. Secondly, we propose algorithms with a new selection mechanism for
  Quality-Diversity algorithms that outperforms all the algorithms
  tested in this paper. Lastly, we present a new collection management that
  overcomes the erosion issues observed when using unstructured
  collections. These three contributions are supported by extensive experimental comparisons of Quality-Diversity algorithms on three
  different experimental scenarios.
\end{abstract}


%
\IEEEpeerreviewmaketitle

\section{Introduction}

Searching for high-quality solutions within a typically high-dimensional search space is an important part of
engineering and research.  Intensive work has been done in recent decades to produce automated procedures to generate these solutions, which are commonly called
``Optimization Algorithms''. The
applications of such algorithms are numerous and range from
modeling purposes to product design~\cite{antoniou2007practical}.
More recently, optimization algorithms have become the core of most
machine learning techniques. For example, they are used to adjust the
weights of neural networks in order to minimize the classification error~\cite{rumelhart1988learning, russell2003artificial}, or to allow robots to
learn new behaviors that maximize their velocity or
accuracy~\cite{cully2015robots, kober2013reinforcement}.

Inspired by the ability of natural evolution to generate species
that are well adapted to their environment, Evolutionary Computation
has a long history in the domain of optimization, particularly in
stochastic optimization~\cite{spall2005introduction}. For example, evolutionary methods have been used to optimize the morphologies and the neural networks of
physical robots~\cite{lipson2000automatic}, and to infer the equations
behind collected data~\cite{schmidt2009distilling}. These optimization
abilities are also the core of Evolutionary Robotics in which
evolutionary algorithms are used to generate neural networks, robot
behaviors, or objects~\cite{eiben2015evolutionary, bongard2006resilient}.

However, from a more general perspective and in contrast with
Artificial Evolution, Natural Evolution does not produce one effective
solution but rather an impressively large set of different organisms,
all well adapted to their respective environment.  Surprisingly, this
divergent search aspect of Natural Evolution is rarely considered in
engineering and research fields, even though the ability to provide a large
and diverse set of high-performing solutions appears to be promising
for multiple reasons.

For example, in a set of effective solutions, each provides an alternative in the
case that one solution turns out to be less effective than expected.
This can happen when the optimization process takes place in
simulation, and the obtained result does not transfer well to
reality (a phenomenon called the reality
gap~\cite{koos2013transferability}).
In this case, a large collection
of solutions can quickly provide a working solution~\cite{cully2015robots}.
Maintaining multiple solutions and using them concurrently to generate actions or predict actions
when done by other agents has also been shown to be very successful in bioinspired motor
control and cognitive robotics experiments\cite{demiris2014information}.

\begin{figure}[!t]
\centering \includegraphics[width=0.75\columnwidth]{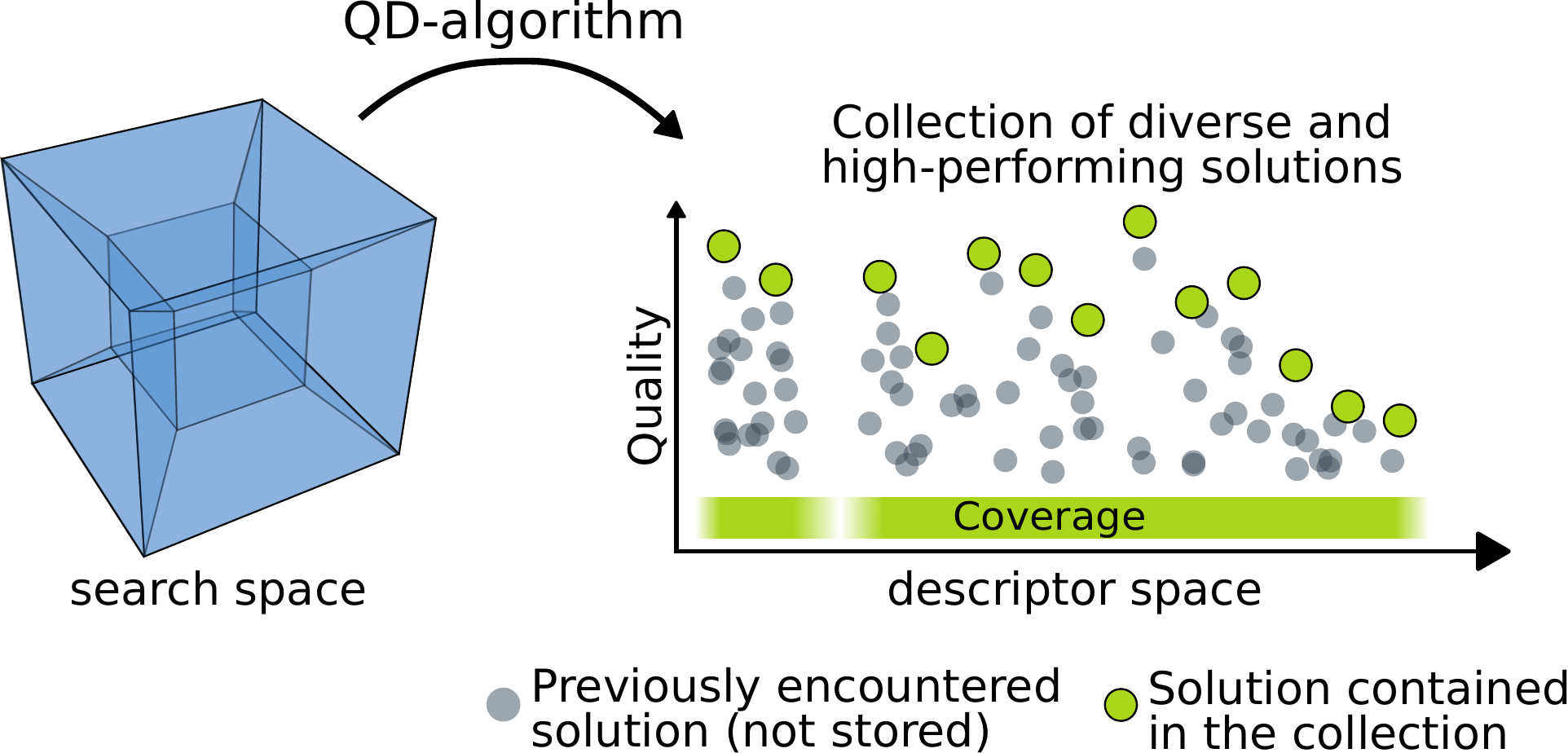}
\caption{The objective of a QD-algorithm is to generate a collection
  of both diverse and high-performing solutions. This collection
  represents a (model free) projection of the high-dimensional search space into a lower dimensional space
  defined by the solution descriptors. The quality of a collection is
  defined by its coverage of the descriptor space and by the global
  quality of the solutions that are kept in the collection.}
\label{fig:concept}
\end{figure}

Moreover, most artificial agents, like robots, should be able to
exhibit different types of behavior in order to accomplish their
mission. For example, a walking robot needs to be able to move not
only forwards, but in every direction and at different speeds, in
order to properly navigate in its environment. Similarly, a robotic
arm needs to be able to reach objects at different locations rather
than at a single, predefined target. Despite this observation, most
optimization techniques that are employed to learn behaviors output
only a single solution: the one which maximizes the optimized
function~\cite{bongard2006resilient, lipson2000automatic,
  kober2013reinforcement}. Learning generic controllers that
are able to solve several tasks is particularly challenging, as it
requires testing each solution on several scenarios to assess their
quality~\cite{cully2013behavioral}. The automatic creation of a
collection of behaviors is likely to overcome these limitations and
will make artificial agents more versatile.

The diversity of the solutions could also be beneficial for the
optimization process itself. The exploration process may find, within the diversity of the solutions, stepping stones that allow
the algorithm to find even higher-performing solutions. Similarly, the
algorithms may be able to solve a given problem faster if they can rely on
solutions that have been designed for different but related situations.
For example, modifying an existing car design to make it lighter might be faster than inventing a completely new design.

Attracted by these different properties several recent works, such as
Novelty Search with Local Competition~\cite{lehman2011evolving} and
the MAP-Elites algorithm~\cite{mouret2015illuminating}, started to
investigate the question of generating large collections of both
\emph{diverse} and \emph{high-performing} solutions. Pugh et
al.~\cite{pugh2015confronting, pugh2016quality} nicely named this
question as \emph{the Quality-Diversity} (QD) challenge.


After a brief description of the origins of QD-algorithms in the next
section, \emph{we unify these algorithms into a single modular framework},
which opens new directions to create QD-algorithms that combine the
advantages of existing methods (see section~\ref{sec:framwork}). Moreover, we introduce a \emph{new QD-algorithm
  based on this framework that outperforms the existing approaches by
  using a new selective pressure, named the ``curiosity score''}. We also introduce a \emph{new archive management approach for
  unstructured archives}, like the novelty archive~\cite{lehman2011abandoning}. The performance of
these contributions is assessed via an extensive experimental
comparison involving numerous variants of QD-algorithms (see
section~\ref{sec:exp}). After the conclusion, we introduce the
open-source library designed for this study, which can be openly used by
interested readers (see section~\ref{sec:lib}).

\section{Related Works and Definitions}\label{sec:BG}
While the notion of Quality-Diversity is relatively recent, the
problem of finding multiple solutions to a problem is a long-standing
challenge.

\subsection{Searching for Local Optima}
This challenge was first addressed by multimodal function
optimization algorithms, including niching methods in Evolutionary
Computation~\cite{goldberg1987genetic,mahfoud1995niching,singh2006comparison},
which aim to find the local optima of a function. These algorithms
mainly involve niche and genotypic diversity preservation
mechanisms~\cite{singh2006comparison}, like
clustering~\cite{yin1993fast} and
clearing~\cite{petrowski1996clearing} methods.

However, in many applications, some interesting solutions are not
captured by the local-optima of the fitness function. For example, it
is important for walking robots to be able to control the walking
speeds, however, there is no guarantee that the performance function
(i.e., the walking speed~\cite{lizotte2008practical, kohl2004policy})
will show local-optima that are diverse enough to provide a complete
range of walking speeds. Typically, if the optimized function is
mono-modal (i.e., without local-optima), the population would tend to
the global-optimum and the diversity of the produced walking behaviors
will not be enough to properly control the robot. For instance, it
will not contain slow behaviors, which are essential for the robot's
manoeuvrability. This example illustrates that sampling the entire
range of possible solutions is not always related to searching for the
local optima, and why it may be useful to have the diversity
preservation mechanism not correlated with the performance function,
but rather based on differences in the solution type.

\subsection{Searching for Diverse Solutions }
Following this idea of a non performance-based diversity mechanism, the Novelty
Search algorithm~\cite{lehman2011abandoning} introduces the idea of
searching for solutions that are different from the previous ones,
without considering their quality. This concept is applied by
optimizing a ``novelty
score'' that characterizes the difference of a solution compared to
those already encountered, which are stored in a ``novelty
archive''. The novelty archive is independent from the population of the evolutionary algorithm. The novelty score is computed as the average distance of
the k-nearest neighboring solutions that currently are in the novelty archive,
while the distances are computed according to a user-defined solution
descriptor (also called a behavioral characterization, or behavioral
descriptor~\cite{lehman2011abandoning, cully2013behavioral}). When the
novelty score of a solution exceeds a pre-defined threshold, this
solution is added to the archive and thus used to compute the novelty score
of future solutions.

The main hypothesis behind this approach is that, in some cases, the
optimal solutions cannot be found by simply maximizing the objective
function. This is because the algorithm first needs to find stepping stones that
are ineffective according to the objective function, but lead to
promising solutions afterwards. A good illustration of this problem
is the ``deceptive maze''~\cite{lehman2011abandoning} in which following the objective function
inevitably leads to a dead-end (a local extremum). The algorithm has
to investigate solutions that lead the agent further from the goal
before being able to find solutions that actually solve the task.

The authors of Novelty Search also introduced the ``Novelty Search
with Local Competition'' algorithm (NSLC)~\cite{lehman2011evolving},
in which the exploration focuses on solutions that are both novel
(according to the novelty score) and locally high-performing. The main
insight consists of comparing the performance of a solution only to
those that are close in the descriptor space. This is achieved with a
``local quality score'' that is defined as the number of the k-nearest
neighboring solutions in the novelty archive with a lower performance
(e.g., slower walking speed~\cite{lehman2011evolving}) than the
considered solution. The exploration is then achieved with a
multi-objective optimization algorithm (e.g.,
NSGA-II~\cite{deb2002fast}) that optimizes both the novelty and local
quality scores of the solutions. However, the local quality score does
not influence the threshold used to select whether an individual is
added to the novelty archive. The final result of NSLC is the
population of the optimization algorithm, which contains solutions
that are both novel and high-performing compared to  other local solutions. In other words, the population gathers solutions that are
both different from those saved in the novelty archive, and
high-performing when compared to similar types of solutions.

The first applications of NSLC consisted of
evolving both the morphology and the behavior of virtual creatures in
order to generate a population containing diverse species, ranging
from slow and massive quadrupeds to fast and lightweight unipedal
hoppers by comparing velocity only between similar species ~\cite{lehman2011evolving}.  In this experiment, the solution
descriptor was defined as the height, the mass and the number of
active joints, while the quality of the solutions was governed by
their walking speed. At the end of the evolutionary process, the
population contained 1,000 different species. These results
represent the very first step in the direction of generating a
collection of diverse and high-performing solutions covering a significant
part of the spectrum of possibilities.

\subsection{Gathering and Improving these Solutions into Collections}
Instead of considering
the population of NSLC as the result of the algorithms, Cully et
al.~\cite{cully2013behavioral} suggested to consider the novelty
archive as the result. Indeed, the aim of the novelty archive is to
keep track of the different solution types that are encountered during
the process, and thus to cover as much as possible of the entire descriptor space. Therefore, the
novelty archive can be considered as a collection of diverse solutions
on its own. However, the solutions are stored in the collection
without considering their quality: as soon as a new type of solution
is found, it is added to archive. While this procedure allows the
archive to cover the entire spectrum of the possible solutions, in the
original version of NSLC only the first encountered solution of each
type is added to the archive. This implies that when finding a better
solution for a solution type already present in the archive, this
solution is not added to the archive. This mechanism prevents the
archive from improving over time.

Based on this observation, a variant of NSLC, named ``Behavioral
Repertoire Evolution''(BR-Evolution~\cite{cully2013behavioral}), has
been introduced to progressively improve the archive's quality by
replacing the solutions that are kept in the archive with better ones
as soon as they are found.  This approach has been applied to generate
``Behavioral Repertoires'' in robotics, which consists of a large
collection of diverse, but effective, behaviors for a robotic agent in
a single run of an evolutionary algorithm.
It has also been used to produce collections of walking
gaits, allowing a virtual six-legged robot to walk in every
direction and at different speeds. The descriptor space is defined as
the final position of the robot after walking for 3 seconds, while
the quality score corresponds to an orientation error.
As we reproduce this experiment in this paper, we provide additional descriptions and technical details in section~\ref{sec:hexa_turn}.

The concepts introduced with BR-Evolution have also later been employed in the Novelty-based
Evolutionary Babbling (Nov-EB)~\cite{maestre2015bootstrapping} that
allows a robot to autonomously discover the possible interactions with
objects in its environment. This work draws a first link between the
QD-algorithms and the domain of developmental robotics, which is also
studied in several other works (see~\cite{benureau2016behavioral} for
overview).

One of the main results that has been demonstrated with BR-Evolution
experiments is that this algorithm is able to generate an effective
collection of behaviors several times faster than by optimizing each
solution independently (at least 5 times faster and about 10 times
more accurate \cite{cully2013behavioral}). By
``recycling'' and improving solutions that are usually discarded by
traditional evolutionary algorithms, the algorithm is able to quickly
find necessary stepping stones. This observation correlates with the earlier
presented hypothesis that QD-algorithms are likely to benefit from the
diversity contained in the collection to improve their optimization and
exploration abilities.

However, it has been noticed that the archive improvement mechanism
may ``erode'' the borders and alter the coverage of the collection~\cite{cully2015evolving}.
Indeed, there are cases where the new, and better, solution found by the
algorithm is less novel than the one it will replace in the
archive. For instance, if high-performance can be more easily achieved for a solution
in the middle of the descriptor space, then it is likely
that the solutions near the borders will progressively be replaced by
slightly better, but less novel, solutions. In addition to eroding
the borders of the collection, this phenomenon will also increase the
density of solutions in regions with a high performance.
For instance, this phenomenon has been observed in the generation of collections
containing different walking and turning
gaits~\cite{cully2015evolving}. The novelty archive of the original
NSLC algorithm had a better coverage of the descriptor space (but with
lower performance scores) than the one from the BR-Evolution, because
it is easier for the algorithms to find solutions that make the robot
walk slowly rather than solutions that make it walk fast or execute
complex turning trajectories (In section \ref{sec:archive} of this paper,
we introduce a new archive management mechanism that overcomes these erosion issues).

\subsection{Evolving the Collection}
Following different inspirations from the works presented above, the Multi-dimensional Archive of
Phenotypic Elites (\emph{MAP-Elites}) algorithm~\cite{mouret2015illuminating}
has been recently introduced. While this algorithm was first designed
to ``illuminate'' the landscape of objective
functions~\cite{clune2013evolutionary}, it showed itself to be an
effective algorithm to generate a collection of solutions that are both
diverse and high-performing. The main difference with NSLC and
BR-Evolution is that, in MAP-Elites, the population of the algorithms
is the collection itself, and the selection, mutations and preservation
mechanisms directly consider the solutions that are stored in the
collection.

In MAP-Elites, the descriptor space is discretized and represented as
a grid. Initially, this grid is empty and the algorithm starts with a
randomly generated set of solutions. After evaluating each solution
and recording its associated descriptor, these solutions are
potentially added to the corresponding grid cells. If the cell is
empty, then the solution is added to the grid, otherwise, only the
best solution among the new one and the one already in the grid is
kept. After the initialization, a solution is randomly selected via a uniform distribution among
those in the grid, and is mutated. The new solution obtained after the
mutation is then evaluated and fitted back in the grid following the
same procedure as in the initialization. This
selection/mutation/evaluation loop is repeated several millions times,
which progressively improves the coverage and the quality of the
collection.

In one of its first applications, MAP-Elites was used to generate
a large collection of different but effective ways to walk in a
straight line by using differently the legs of a six-legged
robot. This collection of behaviors was then used to allow the robot
to quickly adapt to unforeseen damage conditions by selecting a new
walking gait that still works in spite of the
situation~\cite{cully2015robots}. The same algorithm has also been
used to generate behavioral repertoires containing turning
gaits, similarly to the work described previously, and it was shown
that MAP-Elites generates better behavior collections while being
faster than the BR-Evolution algorithm~\cite{cully2015creative}.

The behaviors contained in these collections can be seen as locomotion
primitives and thus can be combined to produce complex behaviors.
Following this idea, the Evolutionary Repertoire-Based Control
(EvoRBC~\cite{duarteevorbc}) evolves a neural network, called the
``arbitrator'', that selects the appropriate behavior in the repertoire,
which was previously generated with MAP-Elites. This approach has
been applied on a four-wheeled steering robot that has to solve a
navigation task through a maze composed of several sharp angles, and a
foraging task in which the robots needs to collect and consume as many
objects as possible.

These applications take advantage of the
non-linear dimensionality reduction provided by MAP-Elites. Indeed,
both applications select behaviors from the descriptor space,
which is composed of fewer than a dozen of dimensions (respectively, 36 to 6 dimensions~\cite{cully2015robots} and 8 to 2 dimensions~\cite{duarteevorbc}), while the
parameter space often consists of several dozen dimensions.
MAP-Elites has been employed in several other applications, including the
generation of different morphologies of soft
robots~\cite{mouret2015illuminating}, or the production of images that
are able to fool deep neural networks~\cite{nguyen2015deep}. It has
also been used to create ``innovation engines'' that are able to
autonomously synthesize pictures that resemble to actual objects
(e.g., television, bagel, strawberry)~\cite{nguyen2015innovation}.

However, the obligation to discretize the descriptor space may be
limiting for some applications, and the uniform random selection may
not be suitable for particularly large collections, as it dilutes the
selection pressure. Indeed, the uniform random selection of
individuals among the collection makes the selection pressure
inversely proportional to the number of solutions actually contained
in the collection. A simple way to mitigate this limitation is to use
a biased selection according to the solution performance or according
to its novelty score (like introduced by Pugh et
al.~\cite{pugh2015confronting, pugh2016quality}).  Another direction
consists in having a number of cells irrespective of the
dimensionality descriptor space, for example by using computational
geometry to uniformly partition the high-dimensional descriptor space
into a pre-defined number of regions~\cite{vassiliades2016scaling}, or
by using Hierarchical Spatial Partitioning~\cite{smith2016rapid}.



\subsection{Quality-Diversity Optimization}
Based on the seminal works presented previously~\cite{lehman2011evolving, mouret2015illuminating, cully2013behavioral} and the formulation of Pugh et al.~\cite{pugh2015confronting, pugh2016quality}, we can outline a
common definition:
\begin{definition}{Quality-Diversity optimization}{def:QD}
A Quality-Diversity optimization algorithm aims to produce a large
collection of solutions that are both as \emph{diverse} and
\emph{high-performing} as possible, which covers a particular
domain, called the descriptor space.
\end{definition}

While this definition is shared with the existing literature,
we also stress the importance of the coverage regularity of the produced
collections. In the vast majority of the applications presented
previously, not only is the coverage of importance but
its uniformity is as well. For example, in the locomotion tasks, an even
coverage of all possible turning abilities of the robot is
required to allow the execution of arbitrary trajectories~\cite{cully2015evolving}.

Based on this definition, the overall performance of a QD-algorithm is
defined by the quality of the produced collection of solutions
according to three criteria:
\begin{enumerate}
\item the coverage of the descriptor space;
\item the uniformity of the coverage; and
\item the performance of the solution found for each type.
\end{enumerate}

\subsection{Understanding the Underlying Mechanisms}
In addition to direct applications, several other works focus on studying the properties of QD-algorithms.
For example, Lehman et al.~\cite{lehman2015enhancing} revealed that
extinction events (i.e., erasing a significant part of the collection)
increases the evolvability of the solutions~\cite{tarapore2015evolvability} and allow the
process to find higher-performing solutions afterwards. For example,
with MAP-Elites, erasing the entire collection except 10 solutions
every 100 000 generations increases the number of filled cells by
$20\%$ and the average quality of the solutions by $50\%$ in some
experimental setups~\cite{lehman2015enhancing}.

In other studies, Pugh et al.~\cite{pugh2015confronting, pugh2016quality}
analyzed the impact of the alignment between the solution descriptor
and the quality score on both Novelty-based approaches (including
NSLC) and MAP-Elites. For example, if the descriptor space represents
the location of the robot in a maze, and the quality score represents the
distance between this position and the exit, then the descriptor space
and the quality score are strongly aligned because the score can be
computed according to the descriptor. The experimental results show
that in the case of such alignments with the quality
score, then novelty-based approaches are more effective than
MAP-Elites, and vice-versa.

Another study also reveals that the choice of the encoding (the mapping between
the genotype and the phenotype) critically impacts the quality
of the produced collections~\cite{tarapore2016different}. The
experimental results link these differences to the locality of the
encoding (i.e., the propensity of the encoding to produce similar
behaviors after a single mutation). In other words, the behavioral
diversity provided by indirect encoding, which is known to empower
traditional evolutionary algorithms~\cite{mouret2012encouraging}, appears to be counterproductive
with MAP-Elites, while the locality of direct encodings allows MAP-Elites
to consistently fill the collection of behaviors.

These different works illustrate the interest of the community in
QD-algorithms and that our understanding of the underlying dynamics is
only in its early stages. However, very few works compare MAP-Elites
and NSLC on the same
applications (the few exceptions being~\cite{pugh2015confronting,pugh2016quality,cully2015creative,smith2016rapid}), or investigate alternative approaches to produce
collections of solutions. One of the goals of this paper is to
introduce a new and common framework for these algorithms to exploit
their synergies and to encourage comparisons and the creation of new
algorithms. The next section introduces this framework.

\section{A united and modular framework for QD-Optimization algorithms}\label{sec:framwork}
\begin{algorithm*}
\small
\caption{QD-Optimization algorithm ( $I$ iterations)}
\label{algo:QD}

\begin{algorithmic}
\State $(\mathcal{A} \leftarrow \emptyset)$\Comment{\emph{Creation of an empty container.}}
\For{iter $  = 1\to I$} \Comment{\emph{The main loop repeats during $I$ iterations.}}
\If{iter $== 1$} \Comment{\emph{Initialization.}}
  \State $\mathbf{b_{parents}}\leftarrow $ random() \Comment{\emph{The first $2$ batches of individuals are generated randomly.}}
  \State $\mathbf{b_{offspring}}\leftarrow $ random() 
\Else \Comment{\emph{The next controllers are generated using the container and/or the previous batch.}}
  \State $\mathbf{b_{parents}}\leftarrow $ selection($\mathcal{A}$, $\mathbf{b_{offspring}}$) \Comment{\emph{Selection of a batch of individuals from the container and/or the previous batch.}}
  \State $\mathbf{b_{offspring}}\leftarrow $ random\_variation($\mathbf{b_{parents}}$) \Comment{\emph{Creation of a randomly modified copy of $\mathbf{b_{parents}}$ (mutation and/or crossover).}}
\EndIf
\ForEach {indiv $\in \mathbf{b_{offspring}}$}
\State $\{\mathbf{desc},\textrm{perf}\}\leftarrow$ eval(indiv) \Comment{\emph{Evaluation of the individual and recording of its descriptor and performance.}}
\If{add\_to\_container(indiv)} \Comment{\emph{``add\_to\_container'' returns true if the individual has been added to the container.}}
\State curiosity(parent(indiv))$+=$ Reward \Comment{\emph{The parent gets a reward by increasing its curiosity score (typically Reward $= 1$).}}
\Else
\State curiosity(parent(indiv))$-=$ Penalty \Comment{\emph{Otherwise, its curiosity score is decreased (typically Penalty $= 0.5$).}}
\EndIf
\EndFor
update\_container() \Comment{\emph{Update of the attributes of all the individuals in the container (e.g. novelty score).}}
\EndFor
\State \Return $\mathcal{A}$
\end{algorithmic}
\end{algorithm*}

As presented in the previous section, most works using or comparing
QD-algorithms consider either MAP-Elites or NSLC-based algorithms, or direct comparisons of these two
algorithms. These comparisons are relevant because of the distinct
origins of these two algorithms. However, they only provide high-level
knowledge and do not provide much insight of properties or
particularities which make one algorithm better than the other.

In this section, we introduce a new and common framework for
QD-algorithms, which can be instantiated with different operators,
such as different selection or aggregation operators, similarly to most
evolutionary algorithms. This framework demonstrates that
MAP-Elites and NSLC can be formulated as the same algorithm using
a different combination of operators. Indeed, specific configurations of
this framework are equivalent to MAP-Elites or NSLC. However, this
framework opens new perspectives as some other configurations lead to
algorithms that share the advantages of both MAP-Elites and NSLC. For
example, it can be used to design an algorithm that is as simple as
MAP-Elites but working on an unstructured archive (rather than a grid), or
to investigate different selection pressures like NSLC. Moreover,
this decomposition of the algorithms allows us to draw conclusions on
the key elements that make an algorithm better than the others (e.g., the selective pressure or the way to form the collection).

This new formulation is composed of two main operators: 1) a
container, which gathers and orders the solutions into a collection,
and 2) the selection operator, which selects the solutions that will
be altered (via mutations and cross-over) during the next batch (or
generation). The selection operator is similar to the selection
operators used in traditional evolutionary algorithms, except that it
considers not only the current population, but all the solutions
contained in the container as well. Other operators can be considered
with this new formulation, like the traditional mutation or cross-over
operators. However, in this paper we only consider the operators described above
that are specific to QD-algorithms.

After a random initialization, the execution of a QD-algorithm
based on this framework follows four steps that are repeated:
\begin{itemize}
\item The selection operator produces a new set of individuals ($\mathbf{b_{parents}}$) that will be altered in order to form the new batch of evaluations ($\mathbf{b_{offspring}}$).
\item The individuals of $\mathbf{b_{offspring}}$ are evaluated and their performance and descriptor are recorded.
\item Each of these individuals is then potentially added to the container, according to the solutions already in the collection.
\item Finally, several scores, like the novelty, the local competition,
  or the curiosity (defined in section \ref{sec:curiosity}) score, are updated.
\end{itemize}
These four steps repeat until a stopping criterion is reached
(typically, a maximum number of iterations) and the algorithm outputs the
collection stored in the container. More details can be found in
the pseudo-code of the algorithm, defined in Algorithm \ref{algo:QD}.
In the following subsections, we will detail different variants of the
container, as well as the selection operators.

\subsection{Containers}
The main purpose of a container is to gather all the solutions
found so far into an ordered collection, in which only the best
\emph{and} most diverse solutions are kept. One of the main
differences between MAP-Elites and NSLC is the way the collection of
solutions is formed. While MAP-Elites relies on an N-dimensional grid,
NSLC uses an unstructured archive based on the Euclidean distance
between solution descriptors. These two different approaches
constitute two different container types. In the following, we will detail their implementation and particularities.

\subsubsection{The Grid}
MAP-Elites employs an N-dimensional grid to form the collection of
solutions~\cite{mouret2015illuminating, cully2015robots}. The
descriptor space is discretized and the different dimensions of the
grid correspond to the dimensions of the solution descriptor. With
this discretization, each cell of the grid represents one solution
type.  In the initial works introducing MAP-Elites, only one solution
can be contained in each cell. However, one can imagine having more
individuals per cell (like in~\cite{pugh2016quality} in which two
individuals are kept). Similarly, in the case of multi-objective
optimization, each cell can represent the Pareto front for each
solution type. Nevertheless, these considerations are outside the
scope of this paper.

\paragraph{Procedure to add solutions into the container}
The procedure to add an individual to the collection is relatively
straight forward. If the cell corresponding to the descriptor of the
individual is empty, then the individual is added to the grid.
Otherwise, if the cell is already occupied, only the individual with the
highest performance is kept in the grid.


\paragraph{Computing the novelty/diversity of a solution}\label{sec:grid_nov}
The inherent structure of the grid provides an efficient way to
compute the novelty of each solution. Instead of considering the
average distance of the k-nearest neighbors as a novelty score, like
suggested in~\cite{lehman2011abandoning}, here we can consider the
number of filled cells around the considered individual. The density
of filled cells of the sub-grid defined around the individual is a
good indicator of the novelty of the solution. Similarly to the ``k''
parameter used in the k-nearest neighbors, the sub-grid is defined
according to a parameter that governs its size, which is defined as
$\pm k$ cells around the individual (in each direction).  In this
case, the score needs to be minimized.

\subsubsection{The Archive}\label{sec:archive}
The novelty archive introduced in the Novelty Search algorithm
consists of an unstructured collection of solutions that are only
organized according to their descriptor and their Euclidean distance. As
introduced in the BR-Evolution algorithm~\cite{cully2013behavioral},
the novelty archive can be used to form the collection of solutions by
substituting solutions when better ones are found. In contrast with
the grid container presented previously, the descriptor
space here is not discretized and the structure of the collection
autonomously emerges from the encountered solutions.

\paragraph{Procedure to add solutions into the container}
The management of the solutions is crucial with this container
because it affects both the quality, and the final coverage of the
collection. A first attempt was proposed in the BR-Evolution
algorithm~\cite{cully2013behavioral} by extending the archive
management of the Novelty Search~\cite{lehman2011abandoning}: an
individual is added to the archive if its novelty score exceeds a
predefined threshold (which can be adjusted over time), or if
it outperforms its nearest neighbor in the archive. In the second
case, the nearest neighbor is removed from the archive and only the
best of the two individuals is kept.

While this archive management is relatively simple, further
experiments reveal underlying limitations~\cite{cully2015evolving}.
First, an individual with the same (or very close) descriptor as
another individual can be added to the archive. Indeed, the novelty
score, which is based on the average distance of the k-nearest
neighbors, can still be relatively high even when two individuals are close if the rest of the collection
is further. One of the consequences of using the novelty score as a
criterion to add the solution in the container is that the collection is
likely to show an uneven density of
solutions~\cite{cully2013behavioral, cully2015evolving}. For example,
experiments in these works show collections that contain a high density of solutions in
certain regions (the inter-individuals distance being notably
lower than the Novelty Score threshold used to add individual into the collection). While this property
can be interesting for some applications, it mainly originates from a side
effect. Second, the same experiments reveal that the replacement
of individuals by better ones can erode the border of the
collection, as discussed in the previous section. Indeed, in some cases, the individuals in the center of
the collection show better performance than the ones in its border
(because of the intrinsic structure of the performance function or because the center has been more intensively explored).
This can lead to the replacement of individuals that are on the border of
the collection by individuals that are closer to the center. This is an
important limitation as it reduces the coverage of the collection, as shown in~\cite{cully2015evolving}.

\begin{figure}
\centering
\includegraphics[width=0.65\columnwidth]{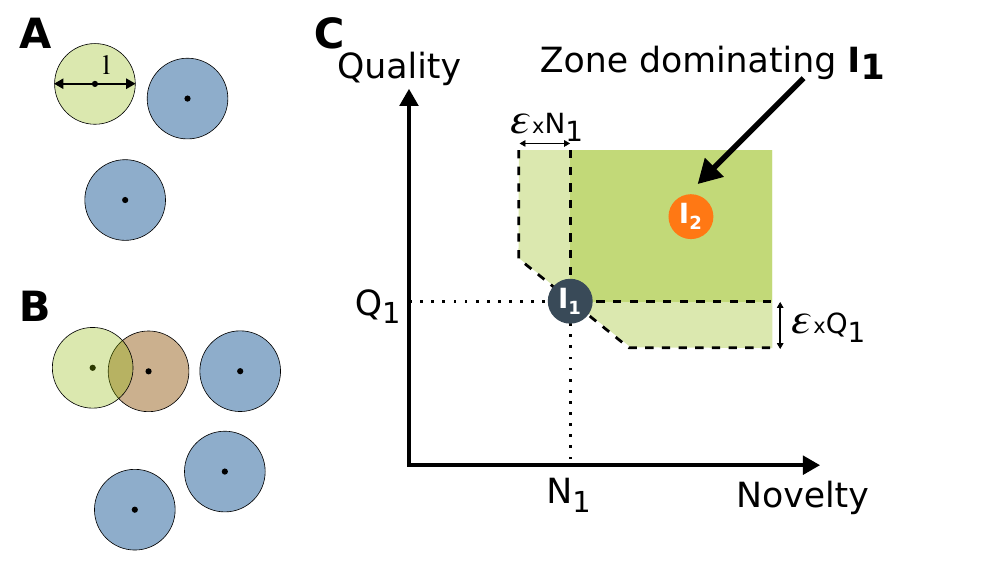}
\caption{Management of collections of solutions based on an
  unstructured archive. A) A solution is directly added to the
  collection if its nearest neighbor from the collection is further
  than $l$. B) Conversely, if the distance is smaller than $l$ (i.e.,
  if the circles overlap), the new solution is not automatically added
  to the collection, but competes against its nearest neighbor. If
  the new solution dominates the one already in the collection, then
  the new solution replaces the previous one. C) In the strict
  $\epsilon$-domination, a solution dominates another one if the
  progress in one objective is larger than the decrease in the other
  objective (up to a predefined value $\epsilon$). }
\label{fig:dominance}
\end{figure}

In order to mitigate these limitations, \emph{we propose the following
new way to manage solutions in the archive}. A solution is added to
the archive if the distance to its nearest neighbor exceeds a
predefined threshold $l$ (see Fig. \ref{fig:dominance}.A). This
parameter defines the maximal density of the collection.
The threshold is similar to the novelty score
threshold used in the original Novelty Search algorithm, except that
in this case we only consider the distance of the nearest neighbor, and not
the average distance of the k-nearest ones.

If the distance between the new individual and its nearest neighbor is
lower than $l$, then this new individual can potentially replace its
nearest neighbor in the collection. This is only the case if its distance from its
second nearest neighbor exceeds the $l$ parameter, such that the
distance among the solutions is preserved (see
Fig. \ref{fig:dominance}.B) and if it improves the overall quality
of the collection. A new individual can improve the overall
collection in two ways: 1) if it has a better quality, which increases
the total quality of the collection or 2) if it has a better novelty
score, which means that it extends the coverage of the
collection. This can be seen as two objectives that need to be
maximized. From this perspective, we can use the definition of
Pareto dominance to decide if an individual should replace another one
already in the collection. Therefore, a simple criterion could be to
replace an existing individual, only if it is dominated by the new
one. However, this criterion is very difficult to reach, as the new
individual should be both better and more diverse than the previous
one. This prevents most new individuals from being added to the
collection, which limits the quality of the produced collections.

In order to soften this criterion, we introduce a variant of the
$\epsilon$-dominance~\cite{laumanns2002combining}, that we name the
\emph{exclusive $\epsilon$-dominance}. In this variant, we tolerate the
dominating individual being worse than the other individual according
to one of the objectives (up to a predefined percentage governed by
$\epsilon$), \emph{only} if it is better on the other objective by \emph{at
  least} the same amount (see Fig. \ref{fig:dominance}.C). This
criterion is more strict than the original $\epsilon$-dominance, which
allows an individual to be dominated by another one that is worse on
both objectives.  From a mathematical point of view, an individual
$x_1$ dominates $x_2$ if these three points are verified:
\begin{enumerate}
\item $N(x_1) >= (1-\epsilon)* N(x_2)$
\item $Q(x_1) >= (1-\epsilon)* Q(x_2)$
\item $(N(x_1)-N(x_2))*Q(x_2)>-(Q(x_1)-Q(x_2))*N(x_2) $
\end{enumerate}
with $N$ corresponding to the Novelty Score and $Q$ to the Quality (or performance) of
an individual, which both need to be maximized\footnote{This
  definition could very likely be generalized to more than two
  objectives, but this question is beyond the scope of this paper.}.
This set of conditions makes the addition of new individuals in the
container more flexible, but rejects individuals that do not improve the
collection.

The experimental results presented in section \ref{sec:exp} demonstrate
that this new archive management overcomes the limitation of the
previous approaches by producing collections with similar coverage and
quality compared with the grid-based container.

\paragraph{Computing the novelty of a solution}
With the archive-based container, the computation of the novelty
score can be done with the traditional approach proposed by Lehman and
Stanley~\cite{lehman2011abandoning}, which consists of the average
distance of the k-nearest neighbors.

\subsubsection{Partial Conclusion}\label{sec:agg_conclusion}
These two different types of containers both present advantages and
disadvantages. On the one hand, the grid-based container provides a
simple and effective way to manage the collection. However, it
requires discretizing the descriptor space beforehand,
which can be problematic for example if the discretization is not
adequate, or needs to be changed over time. On the other hand, the
archive-based container offers more flexibility, as it only requires
the definition of a distance in the descriptor space. For example,
specific distances can be used to compare complex descriptors, like
images, without a strong knowledge of the structure of the descriptor
space (e.g., number of dimensions or limits)~\cite{maestre2015bootstrapping}. However, this advantage is a
disadvantage as well, because it implies that the algorithm needs to find the
appropriate structure of the collection on its own, which represents an
additional challenge compared to the grid-based container.

For these reasons, the choice of the most suitable container
depends more on the considered applications, rather than on their
performance.
Therefore, while we
will consider both of the containers in the experimental section of
this paper, we will not directly compare their results, as the
comparison may not be fair and may be irrelevant with respect to the
considered applications.


These two containers have been designed to provide uniform coverage of
the descriptor space. However, experiments reveal that the
accumulation of density on specific regions of the descriptor space is
a key factor for the Novelty Search algorithm, as it allows the
novelty score to constantly change over time. To avoid this issue, one
can use an additional container in which the density accumulates and
that drives the exploration, while the other container gathers the
collection that will be return to the user. In this paper, we will
only focus on variants using only one container, however we will
consider extending the framework presented in this paper to multiple containers in future
works.

\subsection{Selection Operators}
The second main difference between MAP-Elites
and NSLC is the way the next batch, or population\footnote{We use the word
batch instead of generation because most of the approaches presented in this paper can be used
in a ``steady state'', selecting and evaluating only one
individual at each iteration. However, considering the selection and
evaluation in batches allows the algorithm to execute the evaluation
in parallel, which increases the computational efficiency of the algorithm.}, of solutions is
selected before being evaluated. On the one hand, MAP-Elites forms the
next batch by randomly selecting solutions that are already in the
collection.  On the other hand, NSLC relies on the current population
of solutions and selects the individuals that are both novel and
locally high-performing (according to a Pareto front). This
difference is of significant importance as MAP-Elites uses the entire
collection of solutions, while NSLC only considers a smaller set of
solutions. 

Similarly to the concept of containers, different approaches for
selecting the individuals of the next batch can be considered. In the
following subsections, we will present several selection methods that
can be employed with both container types.

\subsubsection{No Selection}
A naive way to generate the next batch of evaluation is to
generate random solutions. However, this approach is likely
ineffective because it makes the QD-algorithm equivalent to a random
sampling of the search space. In general, this approach provides an
intuition about the difficulty of the task and can be used as a
base-line when comparing alternative approaches.

\subsubsection{Uniform Random Selection}
A second way to select solutions that will be used in
the next batch is to select solutions with a uniform probability from those that are already in the
collection. This approach is the one used in MAP-Elites and follows
the idea that promising solutions are close to each other. In
addition to being relatively simple, this approach has the advantage of being computationally effective.  However, one of its main drawbacks is
that the selection pressure decreases as the number of solutions
in the collection increases (the chance for a solution to be selected
being inversely proportional to the number of solutions in the
collection), which is likely to be ineffective with large collections.

\subsubsection{Score Proportionate Selection}\label{sec:curiosity}
An intuitive way to mitigate the loss of selection pressure from the
random selection is to bias the selection according to a particular
score. Similarly to traditional evolutionary algorithms, the selection
among the solutions of the collection can be biased according to their
quality (fitness), following the roulette wheel or the
tournament-based selection principles~\cite{goldberg1991comparative}.

Other scores can also be considered to bias the selection. For example,
the novelty score of each solution can substitute for the quality score for fostering the algorithm to focus on solutions that are different from the
others.

In addition to these two scores, in this paper we introduce a new score, named the \emph{Curiosity Score}, that
can be used to bias the selection and which is defined as follows:
\begin{definition}{Curiosity Score}{def:curiosity}
The curiosity score represents the propensity of an individual to
generate offspring that are added to the collection.
\end{definition}
A practical implementation (see Algorithm \ref{algo:QD}) consists of
increasing the curiosity score of an individual (initially set to
zero) each time one of its offspring is added to the
collection. Conversely, when an offspring fails to be added to the
archive (because it is not novel or high-performing enough), the
Curiosity Score is decreased. In this paper, we use respectively $1$
and $-0.5$ for the reward and the penalty values. With this
implementation, individuals may gain momentum, but this means that
such individual will be selected more often, making their score more
likely to rapidly decrease.

We named this score ``Curiosity'' because it encourages the algorithm
to focus on individuals that produce interesting solutions, until
nothing new is produced. In other words, the algorithm focuses on
regions of the search space as long as they produce interesting
results, then, when the algorithm gets ``bored'', it focuses its
attention on different regions.  This behavior is similar to the one
of the ``Intelligent Adaptive Curiosity''~\cite{oudeyer2007intrinsic},
while the implementation and the inspiration are strictly different.

A similar approach has recently been introduced to
bias the selection by using the same kind of successful offspring
counter~\cite{lehman2016creative}. The difference is that, in this paper, the counter
is initialized to a fixed value (i.e., 10
in~\cite{lehman2016creative}) instead of starting at 0 like with the
curiosity score, and that when an offspring is added to the
collection, the counter is not incremented (like with the curiosity
score), but rather reset to its maximal value. This difference make
the selection process more forgivable, as only one successful
offspring is enough to make its parent very likely to be selected
again.  While it would be interesting to compare the effect of these
two different, but related, methods, this comparison is out of the
scope of this paper.

Although there is no overall agreement on the definition of
evolvability~\cite{pigliucci2008evolvability}, we can note that our
definition of the curiosity score shares similarities with some of the
first definitions of evolvability, like the one given by Lee Altenberg
who defines the evolvability as ‘‘the ability of a population to
produce variants fitter than any yet
existing’’~\cite{altenberg1994evolution}. One important aspect shared
by these two definitions is that the score or the evolvability may
dynamically change over time according to the state of the population
or the collection, which is rarely considered in
evolvability's definitions. For instance, the definition often used in
Evolutionary Computation ~\cite{tarapore2015evolvability,
  pigliucci2008evolvability, clune2013evolutionary}, which considers
that the evolvability captures the propensity of random variations to
generate phenotypic diversity, depends on the genome of the individual
but not on the state of the population.

%
%

\subsubsection{Population-based Selection}
All selection approaches described so far select the individuals from
the solutions contained in the collection. This represents one of the
main differences introduced by MAP-Elites compared to NSLC and
traditional evolutionary algorithms, as the collection becomes the
``population'' of the algorithm and this population progressively
grows during the evolutionary process. However, to handle the
selection, we can consider employing populations in parallel to the
collection. This is in line with the Novelty Search algorithm which
computes the novelty score based on the Collection (the Novelty
archive), but instead uses a traditional population to handle the
selection.

This approach can be included in the framework proposed in this paper
by initializing the population with the first batch and then, after
each batch evaluation, a new population can be generated based on
the individuals of the current population ($\mathbf{b_{offspring}}$)
and their parents ($\mathbf{b_{parents}}$). Classic selection
approaches, like tournament or score proportionate, can be employed to
select the individuals that will be part of the next population. Like
in the collection-based selection, the selection can be biased
according to either the quality, novelty or curiosity
scores.

\subsubsection{Pareto-based Selection}
The population-based selection approach can be extended to
multi-objective selection, via the Pareto ranking, by taking
inspiration from the NSGA-II algorithm~\cite{deb2002fast}.  In this
paper, we will particularly consider a Pareto-based selection operator
that takes both the novelty score and the local quality score
(number of neighbors that outperform the solution) of the individuals into account.
This selection operator is similar to the selection procedure of NSLC.

\subsubsection{Partial Conclusion}
These different selection operators can all be
equally used with both of the containers presented in the previous
section. While the choice of the container influences the type of
the produced results (e.g., unstructured or discretized descriptor
space, see section \ref{sec:agg_conclusion}), the selection operators will only
influence the quality of the results. Therefore, it is of importance
to know which operators provide the best collection of solutions. In
the following section, we provide a first answer to this question by
comparing the collections produced by the different selection
operators and by investigating their behaviors.

\section{Experimental Comparisons}\label{sec:exp}

\begin{table*}
\centering
\caption{The different combinations of containers and selection operators that are evaluated in this paper. The variants in bold are tested on the three experimental scenarios while the others are only tested on the first one.}
\label{tab:variant_arm}
{\footnotesize
\begin{tabular}{r |cccc}
   Variant name & Container & Selection Op. & Considered Value & Related approach\\
   \hline
   \textbf{arch\_no\_selection} & archive & noselection&-& Random Search / Motor Babbling\\
   \textbf{arch\_random} & archive & random & - & - \\
   \textbf{arch\_pareto} & archive & Pareto & Novelty \& Local Quality&  -\\
   arch\_fitness & archive & Score-based & Fitness & - \\
   arch\_novelty & archive & Score-based & Novelty & MAP-Elites with Novelty~\cite{pugh2015confronting} \\
   \textbf{arch\_curiosity} & archive & Score-based & Curiosity  & - \\
   \textbf{arch\_pop\_fitness} & archive & Population-based & Fitness & Traditional EA\\
   arch\_pop\_novelty & archive & Population-based &Novelty & Novelty Search~\cite{lehman2011abandoning}\\
   arch\_pop\_curiosity & archive & Population-based &Curiosity & - \\
   \textbf{grid\_no\_selection} & grid & noselection&-& Random Search / Motor Babbling\\
   \textbf{grid\_random} & grid & random &-& MAP-Elites~\cite{mouret2015illuminating}\\
   \textbf{grid\_pareto} & grid & Pareto & Novelty \& Local Quality & - \\
   grid\_fitness & grid & Score-based & Fitness & - \\
   grid\_novelty & grid & Score-based &Novelty & -  \\
   \textbf{grid\_curiosity} & grid & Score-based &Curiosity & -  \\
   \textbf{grid\_pop\_fitness} & grid & Population-based & Fitness& Traditional EA\\
   grid\_pop\_novelty & grid & Population-based &Novelty & - \\
   grid\_pop\_curiosity & grid & Population-based &Curiosity & - \\
   \emph{\textbf{NSLC}} &\emph{grid} & \emph{Population \& archive based} &  \emph{Novelty \& Local Quality} & \emph{Novelty Search with Local Competition~\cite{lehman2011evolving}}\\
\end{tabular}}
\end{table*}

To compare the different combinations of containers and selection
operators, we consider three experimental scenarios that take place in
simulation: 1) a highly redundant robotic arm discovering how to reach
points in its vicinity, 2) a virtual six-legged robot learning to walk
in every direction, and 3) the same robot searching for a large number
of ways to walk on a straight line.

In addition to the tested
combinations of containers and selection operators, we include the
original Novelty Search with Local Competition algorithm (NSLC,
\cite{lehman2011evolving}) in our experimental comparisons in order to
assess the influence of the lack of density accumulation in the
descriptor space, as discussed in section
\ref{sec:agg_conclusion}. Like in \cite{pugh2015confronting}, all
individuals of the population are potentially added to a grid
container (the same as the one used with the others variants) after
each generation. We then used the produced grid container to compare
NSLC with the other variants. For this experiment, we used the
implementation of NSLC provided by the Sferes$_{v2}$
framework~\cite{mouret2010sferes}.

In the experiments presented in this paper, we only consider direct
encodings with genomes that are small and fixed in size. It would be
interesting to see how the conclusion drawn from these experiments
hold with large genomes, genomes of increasing complexity over generations, or indirect encodings. For instance, \cite{tarapore2016different}
highlights that indirect encodings may have a negative impact on
QD-algorithms. However, these further considerations are out of the scope of this paper and
will be considered in future works.

\subsection{Quality Metrics}
In order to compare the quality of the collections generated by each
variant, we define four quality metrics that characterize both the
coverage and the performance of the solutions:

\subsubsection{Collection Size} indicates the number of solutions contained in the collection and
thus refers to the proportion of the descriptor space that is
covered by the collection.

\subsubsection{Maximal Quality} corresponds to the quality of the best solution contained in the
collection and indicates if the global extremum of the performance
function has been found (if it is known).

\subsubsection{Total Quality} is the sum of the qualities over all the solutions contained in the
collection. This metric provides information on the global quality of
the collection as it can be improved either by finding additional
individuals or by improving those already in the collection.  It
corresponds to the metric named ``Quality Diversity'' used in~\cite{pugh2015confronting}.

\subsubsection{Total Novelty}
This metric is similar to the previous one, except that the sum
considers the novelty score and not the quality value. This metric
indicates if the collection is well distributed over the description
space or rather if the solutions are highly concentrated. This metric
will not be considered for collections produced with the grid-based
container because the distribution of the solutions is forced by the
grid.

\subsubsection*{Other metrics}
In~\cite{mouret2015illuminating, tarapore2016different}, the authors
presented other metrics to compare collections produced by MAP-Elites.
However, the main idea of these metrics is to normalize the quality of
each solution by the maximal quality that can be achieved by each type
of solution (i.e., by each grid cell). To infer the highest possible
quality for each cell, the authors selected the best solution found by
all the algorithms over all the replicates.  However, this
approach is only feasible with the grid-based container because the
continuous descriptor space used in the archive-based container makes
it challenging to associate and compare each ``solution type''. For this
reason, in this paper we decided to only consider the four metrics
presented previously.

\begin{table}
  \caption{Parameter values used the experiments.}
  \label{tab:params}
        {\footnotesize
          \noindent\begin{tabular}{L{2.1cm}|C{1.6cm}C{1.6cm}C{1.6cm}}
          Parameters& First exp & Second exp &   Third exp\\
          \hline
          Batch size&200 & 200 & 200 \\
          No. of Iterations & 50.000 & 10.000 & 20.000\\
          Descriptor size & 2 & 2 & 6\\
          Genotype size& 8 & 36  & 36 \\
          Genotype type& real values & sampled values & sampled values\\
          Crossover & disabled & disabled &disabled\\
          Mutation rate for each parameter & $12.5\%$ & $5\%$ & $5\%$\\
          Mutation type & Polynomial & Random new value & Random new value\\
                    \hline
          Grid container:&&&\\
          \centering Grid size& $100*100$ & $100*100$ & $5$ cells/dim\\
          \centering Sub-grid depth& $\pm3$ cells & $\pm5$ cells &$\pm1$ cells\\
                    \hline
          Archive container:&&&\\
          \centering l & 0.01 & 0.01 & 0.25\\
          \centering $\epsilon$ & 0.1 & 0.1 & 0.1\\
          \centering k & 15 & 15 & 15 \\
          \hline
          NSLC variant:&&&\\
          \centering $\rho_{\textrm{init}}$ & 0.01 & 0.01 & 1\\
          \centering k & 15 & 15 & 15
          \end{tabular}
}\end{table}

\subsection{The Redundant Arm}\label{sec:arm}
\begin{figure*}[!t]
  \centering
  \includegraphics[width=0.8\linewidth]{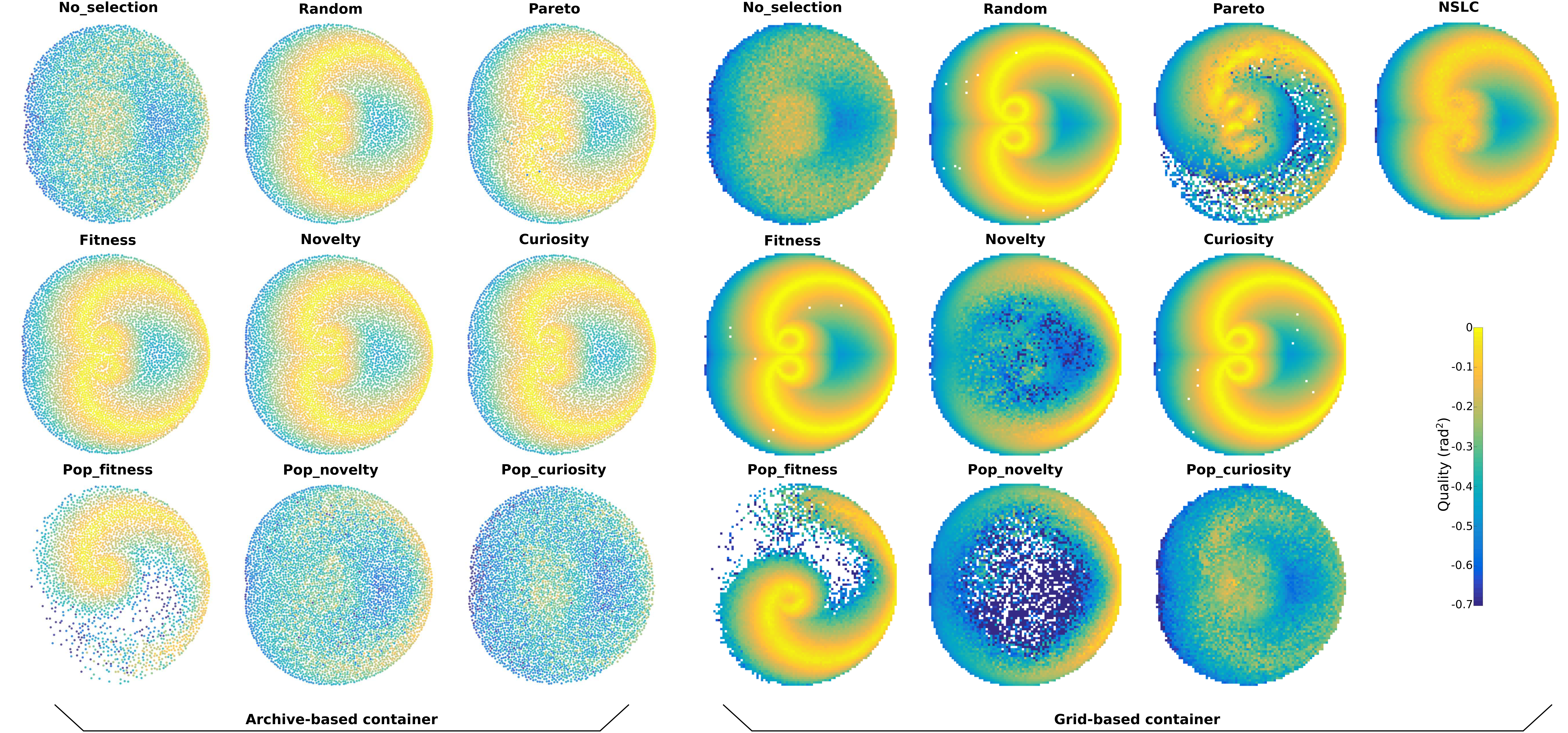}
  \caption{Typical collections of solutions
  produced with QD-algorithms. These collections consist of several thousand colored
  dots or cells that represent the final position of the gripper. The color of each
  dot or cell indicates the quality of the solution (lighter is better).\label{fig:res_arm}}
\end{figure*}

\subsubsection{Experimental Setup}
In this first experimental comparison, we consider a redundant and
planar robotic arm with 8 degrees of freedom that needs to discover
how to reach every point in its vicinity. The quality function
captures the idea that all joints of the arm should contribute equally
to the movement, which allows quick transitions from one configuration
to the next one. This constraint is defined by the variance of the
angular position of the joints when the robot reaches its final configuration, and
needs to be minimized by the
algorithm. This experimental setup illustrates the need of
quality-diversity algorithms because it needs to simultaneously find a
solution for all the reachable positions and to optimize the quality
function for each of them.


To simulate the robotic arm, we consider its kinematic structure,
which provides the location of its gripper according to the angular
position of all joints. The solutions that are optimized by the
algorithms consist of a set of angular positions that govern the final
configuration of the different joints of the robot. Neither the trajectory of
the robot between its initial and final positions, nor internal
collisions are simulated in this experiment.

The solution descriptor is defined as the final position of the
gripper, which is then normalized according to a square bounding box to have
values between 0 and 1. The size of the bounding box is $2*1.1*L$,
where L is the total length of the robot when totally deployed (the
factor $1.1$ is used to leave some room between the border of the
descriptor space and the robot). The center of the box corresponds to
the location of the robot's base.

An extensive set of configurations from the QD-algorithm framework (see
algorithm \ref{algo:QD}) has been tested on this experimental setup
(see Table \ref{tab:variant_arm}), and the execution of each of those
variants has been replicated $20$ times.
The parameter values used for this experiment can be found in Table \ref{tab:params}.

\subsubsection{Results}
\begin{figure*}[!t]
\centering \includegraphics[width=0.95\textwidth]{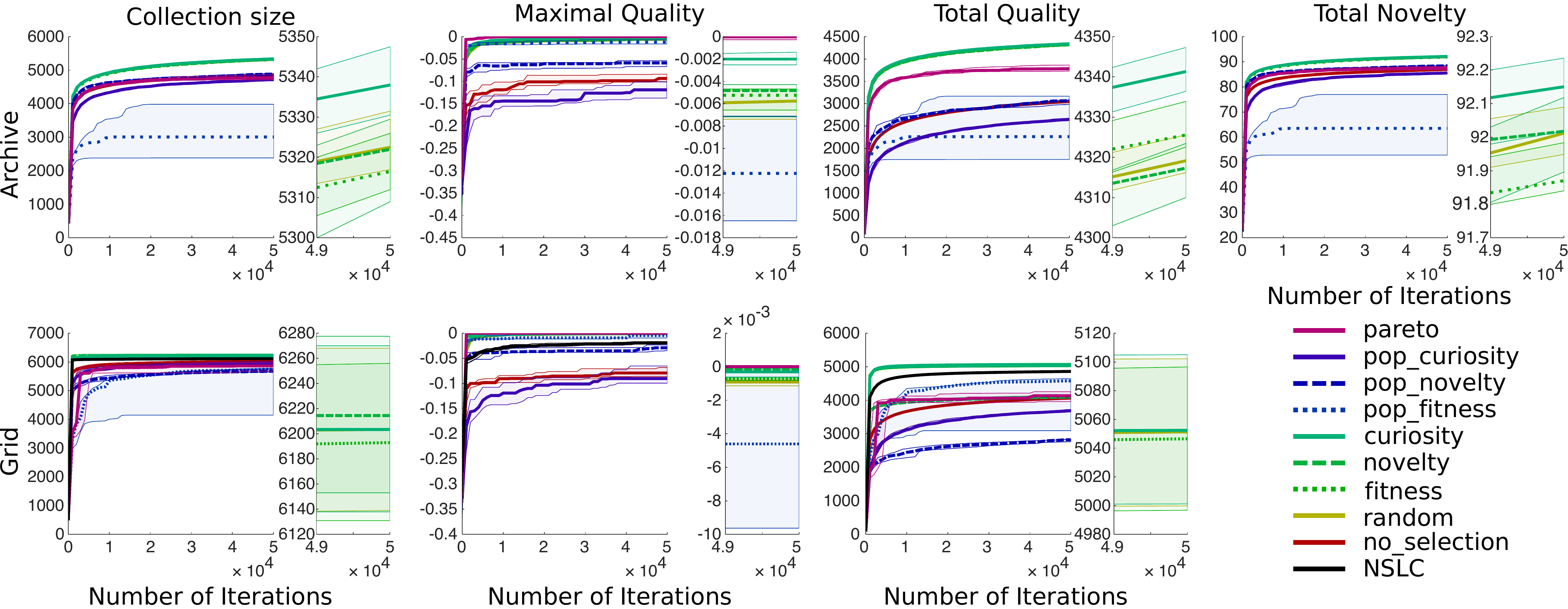}
\caption{Progression of the quality metrics in the redundant arm
  experiment. The first row depicts the results from variants using
  the archive-based container, while the second row considers
  variants with the grid-based container. Because of the difficulty
  to distinguish the different variants, a zoom on the best variants
  during the last 1000 batches is pictured on the right of each
  plot. The middle lines represent the median performance over the 20 replications, while the shaded areas extend to the first and third quartiles. In this experiment, the quality score is negative, thus in order to get a monotonic progression in
  the ``Total Quality'' metric, $+1$ is added to the Quality to have a positive score.}
\label{fig:results}
\end{figure*}

A typical collection of solutions produced by each of the tested
variants is pictured in Figure \ref{fig:res_arm}.  The collections using the archive-based
container appear very similar to those using the other container
type. This similarity, which holds in the other experiments as well,
demonstrates that the archive management introduced in this paper
successfully address the erosion issues described previously.
Theoretically, the ideal result homogeneously covers a quasi-circular region
and the performance (i.e., the color) should be arranged in concentric
shapes resembling cardioids (inverted, heart-shaped curves)\footnote{We can demonstrate that the points with the highest performance are located on a curve resembling a cardioid by computing the position of the end-effector for which all angular positions of the joints are set to
  the same angle (from $-\pi/2$ to $+\pi/2$).}. This
type of collection is found using the random, the fitness or the
curiosity-based selection operators (over the collection) regardless
of the container type used, as well as with the NSLC algorithm. The novelty based selection with the
archive-based container also produces such a collection, while
this is not the case with the grid-based container. It is interesting
to note that the no-selection approach, which can be considered as a
motor babbling or random search, is unable to produce the desired
result. While the coverage is decent, the quality of the gathered
solutions is not satisfactory.

None of the population-based variants managed to produce a collection
that both covers all the reachable space and contains high-performing
solutions. This result could be explained by a convergence of the
population toward specific regions of the collection. Typically, the
population considering the fitness is likely to converge toward regions with high
quality, whereas the population considering the novelty score converges to
the border of the collection. The
results of the variant using a population with the curiosity score could
be explained by the difficulty to keep track of all individuals with
a relatively small population (200 individuals in the population
compared to about 6.000 in the entire collection). The curiosity
score is dynamic, and changes during the evolutionary process (an
individual can have a high curiosity score at one moment, for example
if it reaches a new region of the descriptor space, and 
can have a very low curiosity score later during the process, for
instance when the region becomes filled with good solutions). Therefore, it is likely that the individuals with
the highest curiosity score are not contained in the population.

Moreover, we can observe different results between the grid-based and
the archive-based container variants considering the novelty
score. This difference is likely to originate from the fact that the
novelty score is computed differently in these two container types. Indeed, while in the archive-based container the novelty
score follows the formulation introduced by Lehman and Stanley~\cite{lehman2011abandoning}, in the grid-based container, the novelty
score is computed based on the number of individuals in the
neighboring cells (see section \ref{sec:grid_nov}). Both of these
expressions capture the density of solutions around the considered
individuals. However, in the grid based container, the novelty score
is discretized (because it is related to the number of neighboring
solutions). This discretization is likely to have a strong impact on
score-based selection variants using the novelty score because all
individuals in the center of the collection will have the same and
lowest novelty score (because of all neighboring cells being filled). In the
score-based selection, individuals with the lowest score have nearly no
chance of being selected, which makes the selection focus on the border
of the collection. This behavior is not observed with the
archive-based container because the novelty score is continuous and
the distribution of the solutions in the collection adds some
variability in the novelty score, which makes it impossible to have several
individuals with the lowest novelty score.

While the Pareto-based selection is designed to be similar to the NSLC
algorithm, by keeping in the population individuals that both have a
high novelty and local-competition scores, we can see that the
collection produced by NSLC is significantly better than the
Pareto-based selection approach.  We can explain this poor performance
by the presence of a Pareto-optimal solution in this scenario. Indeed,
the solution in which the robot has all his joint positions set to
zero has the best fitness and is located on the border of the
collection, which provides a high novelty score. It is worth noting that
we can mitigate this issue by implementing a toroidal distance or
container (like in \cite{pugh2016quality}), when such a representation
is compatible with the descriptor space. This is not the case in our
experiments. A behavior that reaches one end of the reachable
space of the robot is not meant to be considered similar to
individuals that reach the opposite end of the reachable space.  For
these reasons, the population is then very likely to converge to this
Pareto-optimal solution and thus, to neglect certain regions of the
collection. The size of the population is probably a limiting factor
as well. A large number of equivalent solutions in terms of
Pareto-dominance exist (all those in the center of the collection with
the highest fitness), which makes it difficult for the population to
cover the entire descriptor space.

NSLC is not impacted in the same way because the original archive
management allows the density to constantly accumulate around
over-explored regions (for instance by varying the novelty threshold,
as described in \cite{lehman2011evolving}). Thanks to this feature,
the novelty score constantly changes over time and makes pareto
optimal solutions disappear quickly. Indeed, the regions that contain
pareto optimal solutions will rapidly see their density increased making the novelty score of the corresponding individuals less
competitive compared with the rest of the population.

It is important to note that the NSLC variant uses two containers and one
population during the evolutionary process. The population and one of the containers
(the novelty archive) are used to drive the exploration process, while
the second container (a grid-based container) gathers the collection
that will be delivered to the user.



The variations of the quality metrics (see Fig. \ref{fig:results})
demonstrate that among all tested variants, the best collections are
provided by variants which perform the selection based on the entire
collection.

The coverage, maximal quality, total quality, and 
total novelty of the collections produced with selection operators
considering the entire collections is higher than those using
population-based selection (all p-values $<7e-8$ from the Wilcoxon rank sum tests\footnote{The reported p-values should be compared to a threshold $\alpha$ (usually set to$0.05$) which is corrected to deal with the ``Multiple Comparisons problem''.  In this paper, all our conclusions about the significance of a difference is given by correcting $\alpha$ according to the Holm-Bonferroni method \cite{shaffer1995multiple}.},
except for the ``(grid/arch)\_pop\_fitness'' approaches which are not significantly different in
terms of maximal quality and for ``grid\_novelty'' which performs
significantly worse than the other collection-based approaches). The only exception is the novelty-based
selection with the grid-based container, which is unable to correctly fill the center of the collection, as it focuses on its borders.

We can note that the variant using the Pareto-based selection with the
archive-based container
produces collections that are better than those from variants using
population-based selection, but worse than those produced by variants
that consider the entire collection for the selection. However, the
Pareto-based selection shows the best results according to the maximal
quality metrics.

While the difference among variants using the entire collection in the
selection with the grid-based container is negligible, the
curiosity-based selection appears to be significantly better (even if the
difference is small) than the other selection approaches on all the
metrics with the archive-based container (all p-values$< 2e-4$ for all
the metrics except for the total novelty in which p-values$<
0.01$). This observation demonstrates that relying on individuals with
a high-propensity to generate individuals that are added to the
collection is a promising selection heuristic.

We can observe that the NSLC variant performs significantly better
than the pareto-based approach and that its
performance is close to, but lower than, those of the variants that use selection
operators considering the entire collections.

\subsection{The Robot Walking in Every Direction}\label{sec:hexa_turn}
\subsubsection{The Experimental Setup}
\begin{figure*}[!t]
\centering \includegraphics[width=0.9\textwidth]{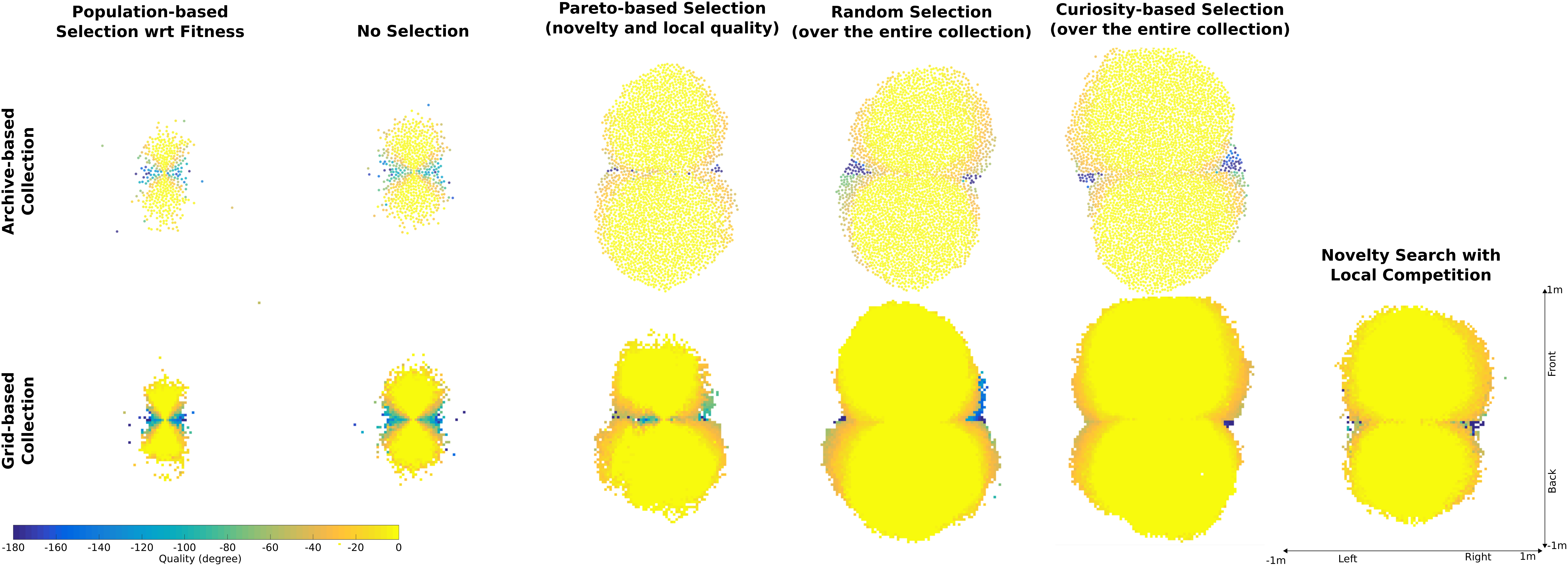}
\caption{Typical collections of solutions
  produced by considered variants in the experiment with the virtual
  legged-robot learning to walk in every direction. The center of each
  collection corresponds to the starting position of the robot and the
  vertical axis represents the front axis of the robot. The position
  of each colored pixel or dot represent the final position of the
  robot after walking for 3 seconds and its color depicts the
  absolute (negative) difference between the robot orientation and the
  desired orientation. Lighter colors indicate better solutions. The collections are symmetrical because the
  robot learns how to walk both forward and backward. This
  possibility, as well as the overall shape of the collection is not
  predefined but rather autonomously discovered by the algorithms.}
\label{fig:res_archives_hexaturn}
\end{figure*}
In this second experimental setup, we consider a six-legged robot in a
physical simulator. The objective of the QD-algorithms is to produce a
collection of behaviors that allows the robot to walk in every
direction and at different speeds. 

This experimental setup has first been introduced
in~\cite{cully2013behavioral}. Each potential solution consists of a
set of 36 parameters (6 per leg) that define the way each of the
robot's joint is moving (the controller is the same as the one used
in~\cite{cully2015robots}). During the evaluation of a solution, the
robot executes the behavior defined by the parameters for three
seconds, and its final position and orientation are recorded. The
descriptor space is defined by the final position of the robot (X and
Y coordinates), while the quality of the solution corresponds to the
orientation error with respect to a desired orientation, which
encourages the robot to follow circular trajectories. These kinds of
trajectories are interesting for planning purposes as any arbitrary
trajectory can be decomposed as a succession of circular arcs. In order to be able to chain circular trajectories, the robot
needs to be aligned with the tangent of these circles at the beginning
and the end of each movement. We can note that only one circular
trajectory goes through both the initial and final positions of the
robot with its tangent aligned with the initial orientation of the
robot. The difference between the final orientation of the robot and
the direction of the tangent of this unique circular trajectory
defines the orientation error, which is minimized by the QD algorithms
(more details can be found in~\cite{cully2013behavioral}).


The usage of the physical simulator makes the experiments
significantly longer (between 4 and 5 hours are required to perform
10,000 batches with one variant). For this reason, the number of
generations has been decreased to 10,000 and only 10 variants (those
in bold in Table \ref{tab:variant_arm}) are considered for this
experiment. This sub-set of variants includes variants that are
related to MAP-Elites, NSLC, Motor Babbling, traditional
population-based EA and the variant considering the curiosity score
over the entire collection.  The execution of each of those
variants has been replicated $10$ times.
The value of the parameters used for this experiment can be found in Table \ref{tab:params}.

\subsubsection{Results}
\begin{figure}[!t]
\centering \includegraphics[width=0.85\columnwidth]{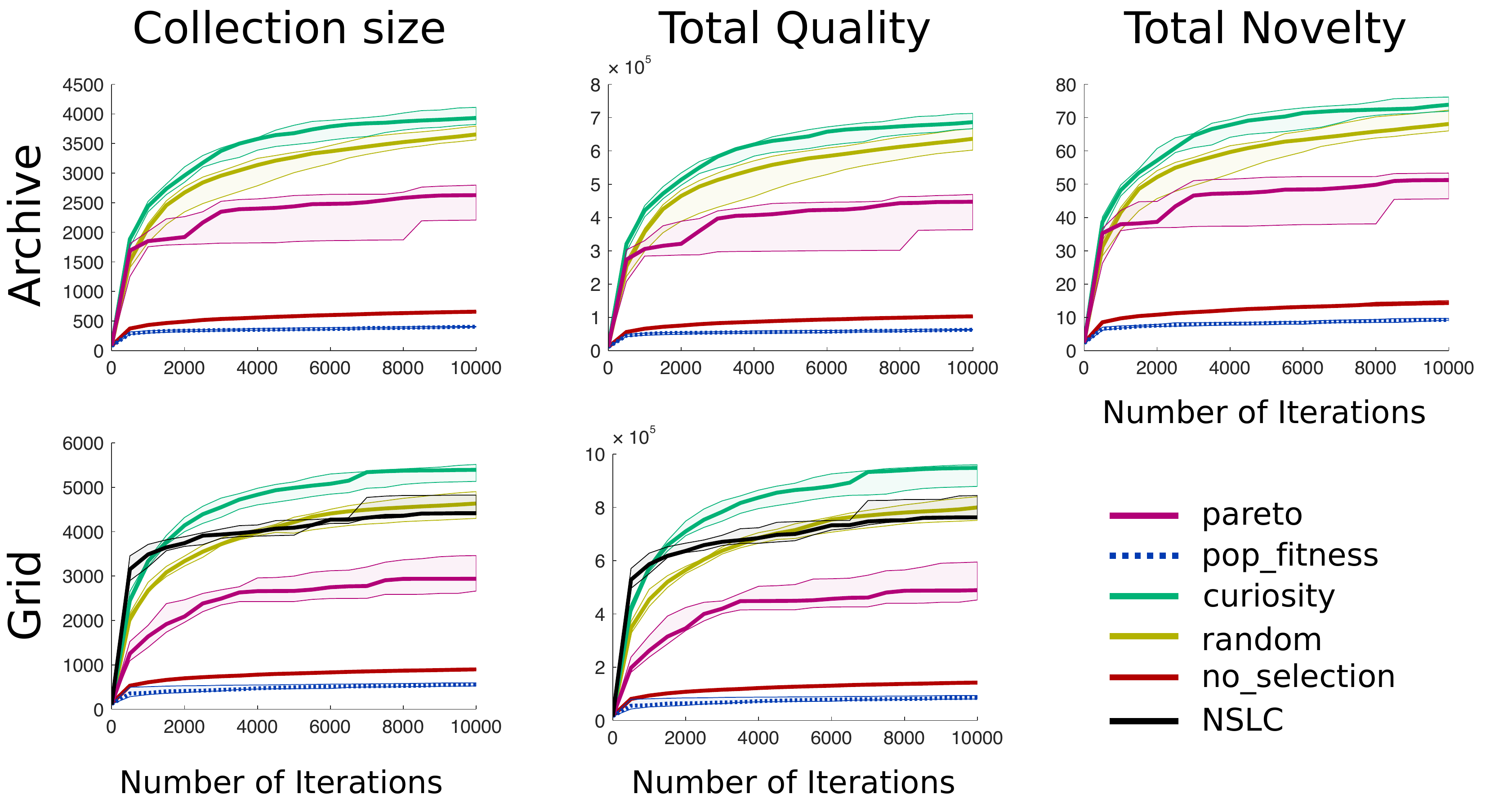}
\caption{Progression of three quality metrics in the turning
  legged-robot experiment. The progression of the maximal quality is
  not depicted because all the variants found at least one solution
  with the highest possible quality (i.e., 0) in fewer than 1.000
  batches. The first row depicts the results from variants using the
  archive-based container, while the second row considers variants
  with the grid-based container. The middle lines represent the median performance over the 10 replications, while the shaded areas extend to the first and third quartiles. In this experiment, the quality
  score is negative, thus in order to get a monotonic progression in
  the ``Total Quality'' metric, $+180$ is added to the Quality to have positive score.}
\label{fig:res_hexaturn}
\end{figure}

From a high-level point of view, the same conclusion as previously
can be drawn based on the resulting collections (see Fig. \ref{fig:res_archives_hexaturn}): The variants
``no\_selection'' and ``pop\_fitness'' produce worse collections than
the other variants, while the variants ``random'', ``curiosity'' and NSLC
generate the best collections. In this experiment, the ``Pareto''
variant performs better than in the previous one. This result can be
explained by the absence of a unique Pareto-optimal solution.

The quality metrics indicate that the ``curiosity'' variants, on both
the grid and the archive containers, significantly outperform the
other algorithms (see Fig. \ref{fig:res_hexaturn}, all
p-values $<0.01$, except when compared to arch\_random in terms of total novelty in which
p-value $=0.05$). These results also demonstrate that this second
experimental scenario is more challenging for the algorithms, as the
difference in the metrics is clear and the performance of the naive
``no\_selection'' is very low.

In this experiment, the NSLC variant shows similar results to the
``random'' variant (which corresponds to the MAP-Elites algorithm). In
particular, the final size of the collection and the final total
quality are not significantly different (p-values$<0.61$). However,
the performance of the ``curiosity'' approach remains significantly better
on both aspects (p-values$<0.0047$) compared to NSLC.

\subsection{The Robot Walking with Different Gaits }
\subsubsection{The Experimental Setup}
In this third experimental setup, we use the same virtual robot as in
the previous experiment with the same controller. However, in this case the robot has to learn a large collection of gaits to walk
 in a straight line as fast as possible. This scenario is inspired by~\cite{cully2015robots}.

In this experiment, the quality score is the traveled distance after
walking for 3 seconds, and the solution descriptor is the proportion
of time that each leg is in contact with the ground. The descriptor
space has thus 6 dimensions in this experiment. The experiment has been replicated 10 times and the other parameters of
the algorithm can be found in Table \ref{tab:params}.

\subsubsection{Results}
\begin{figure}[!t]
\centering \includegraphics[width=0.85\columnwidth]{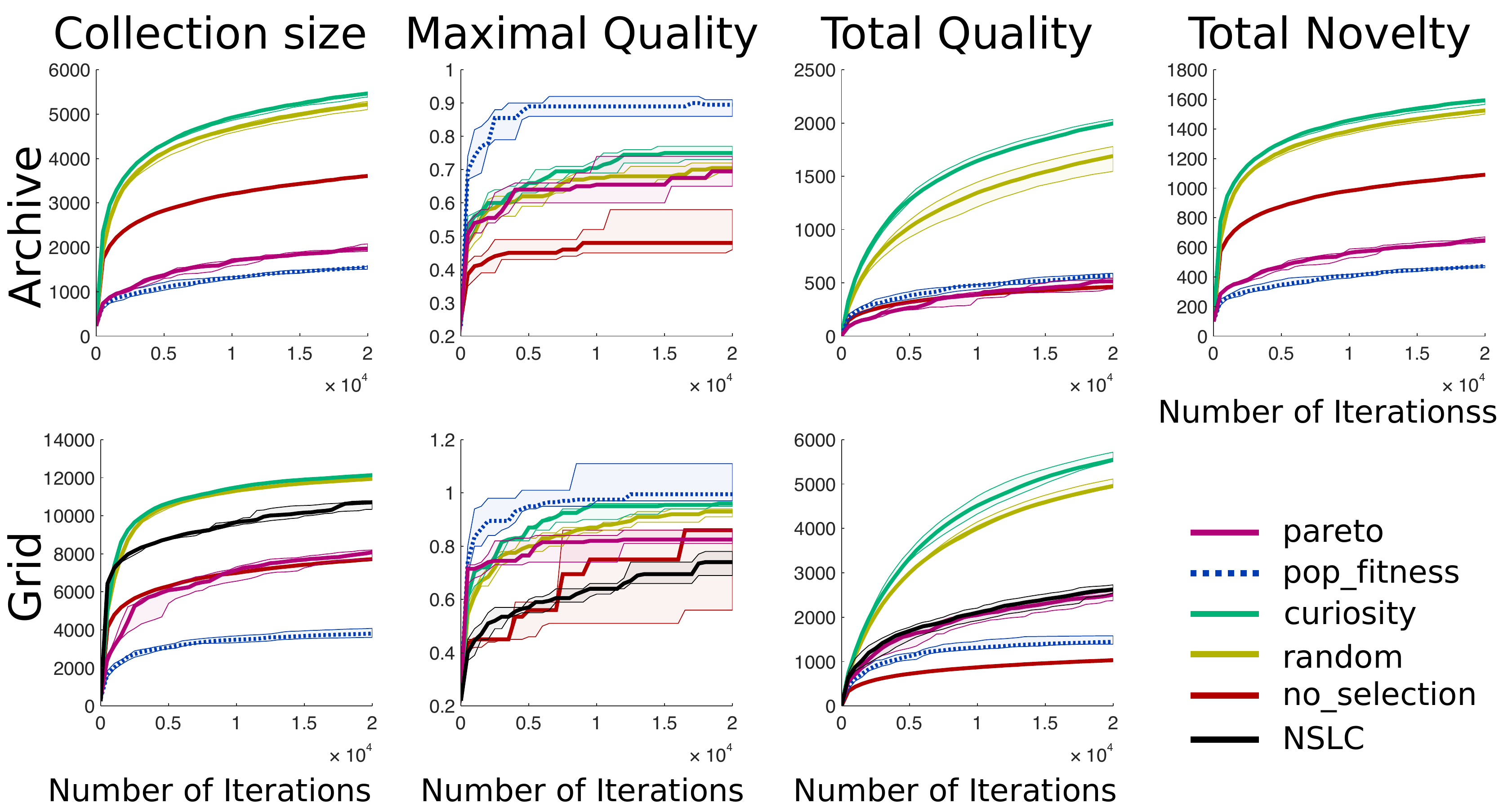}
\caption{Progression of the four quality metrics in the experiment
  with the legged-robot learning different ways to walk in a straight
  line. The first row depicts the results from variants using the
  archive-based container, while the second row considers variants
  with the grid-based container. The middle lines represent the median performance over the 10 replications, while the shaded areas extend to the first and third quartiles.}
\label{fig:res_hexawalk}
\end{figure}

From a general point of view, the same conclusion as in the previous
experiments can be drawn from the progression of quality metrics (see
Fig.\ref{fig:res_hexawalk})\footnote{Visualizations of the collections
  are not provided in this experiment because of the
  high-dimensionality of the descriptor-space. While the grid-based
  collections could have been depicted with the same approach as in~\cite{cully2015robots}, this approach cannot be applied with the
  archive-based container.}. Variants selecting individuals from the
whole collection significantly outperform, in terms of coverage,
total quality and diversity, those that consider populations (all the p-values$<2e-4$). In
particular, the curiosity-based selection operator shows the best
results both with the grid-based and the archive-based containers.
For instance, one can note that the total quality achieved by the
random selection (second best approach) after 20,000 batches, is achieved by the curiosity-based selection after only 11,000
batches with the archive-based container and 13,500 batches with the
grid-based container.

In contrast with the previous experiment, the ``no\_selection''
variants manage to achieve good coverage (about half
of the coverage produced by the variants using the collection-wise
selection). However, they show the worst results according to the total quality and
the maximal quality metrics.

The variants using the population-based selection with respect to the
performance show the opposite results. While the coverage of this
variant is the worst among all the evaluated variants with both of the
container types, this selection approach, which is similar to a
traditional EA, found the solutions with the best quality (the fastest
way to walk). In particular, the performance achieved with this variant
significantly outperforms the best solutions compared to every other
variant, even those using the collection-wise selection (p-values$<0.0017$). This observation shows that the best variants tested so far
are not always able to find the global extremum of the quality. The
quality difference between the ``pop\_fitness'' variants and the
others is smaller with the grid-based container than with the
archive-based. This quality difference could be explained by the
difference in the collection sizes, or the additional difficulty of
finding the inherent structure of the collection for the archive-based
container.

The Pareto-based variants are low-performing in this experiment. They
show neither a good coverage (similar to the ``no\_selection'' or the
``pop\_fitness'' variants) nor a good maximal quality (lower than the
variants with a collection-wise selection). It is difficult to
understand the reasons for such a low performance in this experiment,
as the behavioral space is 6 dimensional, making it hard to
visualize. However, it is likely that it happens for the same reasons
as in the previous experiments, like a premature convergence to the
border of the collection (which show relatively bad performance), or
the existence of a Pareto-optimal solution.  In contrast with the
Pareto-based variants, NSLC achieves good coverage of the behavioral
space in this experiment, while smaller than the ``random'' and
``curiosity'' ones. However, the maximal quality found on the produced
collection is lower than most of the considered variants
(p-values$<0.0374$\footnote{These p-values do not reject the null-hypothesis based on the Holm-Bonferroni method with a $\alpha=0.05$, but reject it with $\alpha=0.1$.} except with the ``no\_selection'' variant,
p-value$=0.9696$), and the global quality of the collections is
equivalent to those of the Pareto-based variant.

\section{Conclusion and Discussion}
In this paper, we presented three new contributions. First, we
introduced a new framework that unifies QD-algorithms, showing for
example that MAP-Elites and the Novelty Search with Local Competition
are two different configurations of the same algorithm.  Second, we
suggested a new archive management procedure that copes with the
erosion issues observed with the previous approaches using
unstructured archives (like BR-evolution). This new procedure
demonstrates good results as it
allows the algorithms to produce unstructured collections with the same coverage as those with grid
containers, which was not the case with the previous management
procedure~\cite{cully2015creative}. Finally, we proposed a new
selective pressure specific for QD-algorithms, named ``curiosity
score'' that shows very promising results by outperforming all the
existing QD-algorithms on all the experiments presented in this paper.

In addition to these three contributions, we presented 
the results of an experimental comparison between a large number of
QD-algorithms, including MAP-Elites and NSLC. One of the main results
that can be outlined from these experiments is that selection
operators considering the collection instead of a population showed
better performance on all scenarios.
We can hypothesize that this results from the inherent diversity of solutions contained in the
collection. Indeed, several works suggest that maintaining the behavioral
diversity in populations of evolutionary algorithms (via additional objective for
example) is a key factor to avoid local extremum and to find promising
stepping stones~\cite{mouret2012encouraging,
  lehman2011abandoning}.

Another fundamental lesson learned from the experiments presented in this
paper is about the importance of allowing the density of solutions to
increase in diverse regions of the archive to obtain the full
effectiveness the NSLC. This can be achieved by varying the
novelty-score threshold or via probabilistic addition to the
archive\cite{lehman2015enhancing}. While such mechanisms are often
used in the literature, their importance is rarely highlighted by
experimental comparisons like in this paper. In particular, we demonstrated that algorithms
using the novelty score, but with archives in which the density does
not increase, are unable to show similar results to NSLC, because they are severely
impacted by certain aspects of the fitness landscape (e.g., presence of Pareto-optimal
solutions). 

This unified and modular framework for QD-algorithms is intended to
encourage new research directions via novel container types, selection
operators, or selective pressures that are specific to this domain. We
expect that the emergence of new QD-algorithms will provide insights
about the key factors for producing the best collection of solutions.

\section{Quality Diversity Library}\label{sec:lib}
The source code of the QD-algorithm framework is available at
\url{https://github.com/sferes2/modular_QD}. It is based on the Sferes$_{v2}$
framework~\cite{mouret2010sferes} and implements both the grid-based
and archive-based containers and several selection operators, including all
those that have been evaluated in this paper. The source code of the
experimental setups is available at the same location and can be used by
interested readers to investigate and evaluate new QD-algorithms.

The implementation allows researchers to easily
implement and evaluate new combinations of operators, while maintaining
high execution speed. For this reason, we followed the policy-based
design in C++~\cite{alexandrescu2001modern}, which allows
developers to replace the behavior of the program simply by changing
the template declarations of the algorithm. For example, changing
from the grid-based container to the archive-based one only requires
changing ``container::Grid'' to ``container::Archive'' in the
template definition of the QD-algorithm object. Moreover, the
modularity provided by this design pattern does not add any overhead,
contrary to classic Object-Oriented Programming design. Interested
readers are welcome to use and to contribute to the source code.

\section*{Acknowledgment}
This work was supported by the EU Horizon2020 project PAL
(643783-RIA). The authors gratefully acknowledge the support from the
members of the Personal Robotics Lab.

\ifCLASSOPTIONcaptionsoff
  \newpage
\fi



\bibliographystyle{IEEEtran}
\bibliography{IEEEabrv,biblio}

\begin{thebibliography}{10}
\providecommand{\url}[1]{#1}
\csname url@samestyle\endcsname
\providecommand{\newblock}{\relax}
\providecommand{\bibinfo}[2]{#2}
\providecommand{\BIBentrySTDinterwordspacing}{\spaceskip=0pt\relax}
\providecommand{\BIBentryALTinterwordstretchfactor}{4}
\providecommand{\BIBentryALTinterwordspacing}{\spaceskip=\fontdimen2\font plus
\BIBentryALTinterwordstretchfactor\fontdimen3\font minus
  \fontdimen4\font\relax}
\providecommand{\BIBforeignlanguage}[2]{{%
\expandafter\ifx\csname l@#1\endcsname\relax
\typeout{** WARNING: IEEEtran.bst: No hyphenation pattern has been}%
\typeout{** loaded for the language `#1'. Using the pattern for}%
\typeout{** the default language instead.}%
\else
\language=\csname l@#1\endcsname
\fi
#2}}
\providecommand{\BIBdecl}{\relax}
\BIBdecl

\bibitem{antoniou2007practical}
A.~Antoniou and W.-S. Lu, \emph{Practical optimization: algorithms and
  engineering applications}.\hskip 1em plus 0.5em minus 0.4em\relax Springer
  Science \& Business Media, 2007.

\bibitem{rumelhart1988learning}
D.~E. Rumelhart, G.~E. Hinton, and R.~J. Williams, ``Learning representations
  by back-propagating errors,'' \emph{Cognitive modeling}, vol.~5, no.~3, p.~1,
  1988.

\bibitem{russell2003artificial}
S.~J. Russell and P.~Norvig, \emph{Artificial intelligence: a modern approach},
  2003.

\bibitem{cully2015robots}
A.~Cully, J.~Clune, D.~Tarapore, and J.-B. Mouret, ``Robots that can adapt like
  animals,'' \emph{Nature}, vol. 521, no. 7553, pp. 503--507, 2015.

\bibitem{kober2013reinforcement}
J.~Kober, J.~A. Bagnell, and J.~Peters, ``Reinforcement learning in robotics: A
  survey,'' \emph{International Journal of Robotics Research}, vol.~32, no.~11,
  p. 1238, 2013.

\bibitem{spall2005introduction}
J.~C. Spall, \emph{Introduction to stochastic search and optimization:
  estimation, simulation, and control}.\hskip 1em plus 0.5em minus 0.4em\relax
  John Wiley \& Sons, 2005, vol.~65.

\bibitem{lipson2000automatic}
H.~Lipson and J.~B. Pollack, ``Automatic design and manufacture of robotic
  lifeforms,'' \emph{Nature}, vol. 406, no. 6799, pp. 974--978, 2000.

\bibitem{schmidt2009distilling}
M.~Schmidt and H.~Lipson, ``Distilling free-form natural laws from experimental
  data,'' \emph{science}, vol. 324, no. 5923, pp. 81--85, 2009.

\bibitem{eiben2015evolutionary}
A.~E. Eiben and J.~Smith, ``From evolutionary computation to the evolution of
  things,'' \emph{Nature}, vol. 521, no. 7553, pp. 476--482, 2015.

\bibitem{bongard2006resilient}
J.~Bongard, V.~Zykov, and H.~Lipson, ``Resilient machines through continuous
  self-modeling,'' \emph{Science}, vol. 314, no. 5802, 2006.

\bibitem{koos2013transferability}
S.~Koos, J.-B. Mouret, and S.~Doncieux, ``The transferability approach:
  Crossing the reality gap in evolutionary robotics,'' \emph{Evolutionary
  Computation, IEEE Transactions on}, vol.~17, no.~1, pp. 122--145, 2013.

\bibitem{demiris2014information}
Y.~Demiris, L.~Aziz-Zadeh, and J.~Bonaiuto, ``Information processing in the
  mirror neuron system in primates and machines,'' \emph{Neuroinformatics},
  vol.~12, no.~1, pp. 63--91, 2014.

\bibitem{cully2013behavioral}
A.~Cully and J.-B. Mouret, ``Behavioral repertoire learning in robotics,'' in
  \emph{Proceedings of the 15th annual conference on Genetic and Evolutionary
  Computation}.\hskip 1em plus 0.5em minus 0.4em\relax ACM, 2013, pp. 175--182.

\bibitem{lehman2011evolving}
J.~Lehman and K.~O. Stanley, ``Evolving a diversity of virtual creatures
  through novelty search and local competition,'' in \emph{Proceedings of the
  13th annual conference on Genetic and Evolutionary Computation}.\hskip 1em
  plus 0.5em minus 0.4em\relax ACM, 2011, pp. 211--218.

\bibitem{mouret2015illuminating}
J.-B. Mouret and J.~Clune, ``Illuminating search spaces by mapping elites,''
  \emph{arXiv preprint arXiv:1504.04909}, 2015.

\bibitem{pugh2015confronting}
J.~K. Pugh, L.~Soros, P.~A. Szerlip, and K.~O. Stanley, ``Confronting the
  challenge of quality diversity,'' in \emph{Proceedings of the 2015 on Genetic
  and Evolutionary Computation Conference}.\hskip 1em plus 0.5em minus
  0.4em\relax ACM, 2015, pp. 967--974.

\bibitem{pugh2016quality}
J.~K. Pugh, L.~B. Soros, and K.~O. Stanley, ``Quality diversity: A new frontier
  for evolutionary computation,'' \emph{Frontiers in Robotics and AI}, vol.~3,
  p.~40, 2016.

\bibitem{lehman2011abandoning}
J.~Lehman and K.~O. Stanley, ``Abandoning objectives: Evolution through the
  search for novelty alone,'' \emph{Evolutionary Computation}, vol.~19, no.~2,
  pp. 189--223, 2011.

\bibitem{goldberg1987genetic}
D.~E. Goldberg and J.~Richardson, ``Genetic algorithms with sharing for
  multimodal function optimization,'' in \emph{Genetic algorithms and their
  applications: Proceedings of the Second International Conference on Genetic
  Algorithms}.\hskip 1em plus 0.5em minus 0.4em\relax Hillsdale, NJ: Lawrence
  Erlbaum, 1987, pp. 41--49.

\bibitem{mahfoud1995niching}
S.~W. Mahfoud, ``Niching methods for genetic algorithms,'' \emph{Urbana},
  vol.~51, no. 95001, pp. 62--94, 1995.

\bibitem{singh2006comparison}
G.~Singh and K.~Deb~Dr, ``Comparison of multi-modal optimization algorithms
  based on evolutionary algorithms,'' in \emph{Proceedings of the 8th annual
  conference on Genetic and Evolutionary Computation}.\hskip 1em plus 0.5em
  minus 0.4em\relax ACM, 2006, pp. 1305--1312.

\bibitem{yin1993fast}
X.~Yin and N.~Germay, ``A fast genetic algorithm with sharing scheme using
  cluster analysis methods in multimodal function optimization,'' in
  \emph{Artificial neural nets and genetic algorithms}.\hskip 1em plus 0.5em
  minus 0.4em\relax Springer, 1993.

\bibitem{petrowski1996clearing}
A.~P{\'e}trowski, ``A clearing procedure as a niching method for genetic
  algorithms,'' in \emph{Evolutionary Computation, 1996., Proceedings of IEEE
  International Conference on}.\hskip 1em plus 0.5em minus 0.4em\relax IEEE,
  1996, pp. 798--803.

\bibitem{lizotte2008practical}
D.~J. Lizotte, \emph{Practical bayesian optimization}.\hskip 1em plus 0.5em
  minus 0.4em\relax University of Alberta, 2008.

\bibitem{kohl2004policy}
N.~Kohl and P.~Stone, ``Policy gradient reinforcement learning for fast
  quadrupedal locomotion,'' in \emph{Proceedings of the IEEE International
  Conference on Robotics and Automation (ICRA)}, vol.~3.\hskip 1em plus 0.5em
  minus 0.4em\relax IEEE, 2004, pp. 2619--2624.

\bibitem{deb2002fast}
K.~Deb, A.~Pratap, S.~Agarwal, and T.~Meyarivan, ``A fast and elitist
  multiobjective genetic algorithm: Nsga-ii,'' \emph{Evolutionary Computation,
  IEEE Transactions on}, vol.~6, no.~2, pp. 182--197, 2002.

\bibitem{maestre2015bootstrapping}
C.~Maestre, A.~Cully, C.~Gonzales, and S.~Doncieux, ``Bootstrapping
  interactions with objects from raw sensorimotor data: a novelty search based
  approach,'' in \emph{Development and Learning and Epigenetic Robotics
  (ICDL-EpiRob), 2015 Joint IEEE International Conference on}.\hskip 1em plus
  0.5em minus 0.4em\relax IEEE, 2015, pp. 7--12.

\bibitem{benureau2016behavioral}
F.~Benureau and P.-Y. Oudeyer, ``Behavioral diversity generation in autonomous
  exploration through reuse of past experience,'' \emph{Frontiers in Robotics
  and AI}, vol.~3, p.~8, 2016.

\bibitem{cully2015evolving}
A.~Cully and J.-B. Mouret, ``Evolving a behavioral repertoire for a walking
  robot,'' \emph{Evolutionary Computation}, 2015.

\bibitem{clune2013evolutionary}
J.~Clune, J.-B. Mouret, and H.~Lipson, ``The evolutionary origins of
  modularity,'' \emph{Proceedings of the Royal Society of London B: Biological
  Sciences}, vol. 280, no. 1755, 2013.

\bibitem{cully2015creative}
A.~Cully, ``Creative adaptation through learning,'' Ph.D. dissertation,
  Universit{\'e} Pierre et Marie Curie, 2015.

\bibitem{duarteevorbc}
M.~Duarte, J.~Gomes, S.~M. Oliveira, and A.~L. Christensen, ``Evorbc:
  Evolutionary repertoire-based control for robots with arbitrary locomotion
  complexity,'' in \emph{Proceedings of the 25th annual conference on Genetic
  and Evolutionary Computation}.\hskip 1em plus 0.5em minus 0.4em\relax ACM,
  2016.

\bibitem{nguyen2015deep}
A.~Nguyen, J.~Yosinski, and J.~Clune, ``Deep neural networks are easily fooled:
  High confidence predictions for unrecognizable images,'' in \emph{Conference
  on Computer Vision and Pattern Recognition}.\hskip 1em plus 0.5em minus
  0.4em\relax IEEE, 2015.

\bibitem{nguyen2015innovation}
A.~M. Nguyen, J.~Yosinski, and J.~Clune, ``Innovation engines: Automated
  creativity and improved stochastic optimization via deep learning,'' in
  \emph{Proceedings of the 2015 Annual Conference on Genetic and Evolutionary
  Computation}.\hskip 1em plus 0.5em minus 0.4em\relax ACM, 2015, pp. 959--966.

\bibitem{vassiliades2016scaling}
V.~Vassiliades, K.~Chatzilygeroudis, and J.-B. Mouret, ``Scaling up map-elites
  using centroidal voronoi tessellations,'' \emph{arXiv preprint
  arXiv:1610.05729}, 2016.

\bibitem{smith2016rapid}
D.~Smith, L.~Tokarchuk, and G.~Wiggins, ``Rapid phenotypic landscape
  exploration through hierarchical spatial partitioning,'' in
  \emph{International Conference on Parallel Problem Solving from
  Nature}.\hskip 1em plus 0.5em minus 0.4em\relax Springer, 2016, pp. 911--920.

\bibitem{lehman2015enhancing}
J.~Lehman and R.~Miikkulainen, ``Enhancing divergent search through extinction
  events,'' in \emph{Proceedings of the 2015 on Genetic and Evolutionary
  Computation Conference}.\hskip 1em plus 0.5em minus 0.4em\relax ACM, 2015,
  pp. 951--958.

\bibitem{tarapore2015evolvability}
D.~Tarapore and J.-B. Mouret, ``Evolvability signatures of generative
  encodings: beyond standard performance benchmarks,'' \emph{Information
  Sciences}, vol. 313, pp. 43--61, 2015.

\bibitem{tarapore2016different}
D.~Tarapore, J.~Clune, A.~Cully, and J.-B. Mouret, ``How do different encodings
  influence the performance of the map-elites algorithm?'' in \emph{Genetic and
  Evolutionary Computation Conference}, 2016.

\bibitem{mouret2012encouraging}
J.-B. Mouret and S.~Doncieux, ``Encouraging behavioral diversity in
  evolutionary robotics: An empirical study,'' \emph{Evolutionary Computation},
  vol.~20, no.~1, pp. 91--133, 2012.

\bibitem{laumanns2002combining}
M.~Laumanns, L.~Thiele, K.~Deb, and E.~Zitzler, ``Combining convergence and
  diversity in evolutionary multiobjective optimization,'' \emph{Evolutionary
  Computation}, vol.~10, no.~3, pp. 263--282, 2002.

\bibitem{goldberg1991comparative}
D.~E. Goldberg and K.~Deb, ``A comparative analysis of selection schemes used
  in genetic algorithms,'' \emph{Foundations of genetic algorithms}, 1991.

\bibitem{oudeyer2007intrinsic}
P.-Y. Oudeyer, F.~Kaplan, and V.~V. Hafner, ``Intrinsic motivation systems for
  autonomous mental development,'' \emph{Evolutionary Computation, IEEE
  Transactions on}, vol.~11, no.~2, pp. 265--286, 2007.

\bibitem{lehman2016creative}
J.~Lehman, S.~Risi, and J.~Clune, ``Creative generation of 3d objects with deep
  learning and innovation engines,'' in \emph{Proceedings of the 7th
  International Conference on Computational Creativity}, 2016.

\bibitem{pigliucci2008evolvability}
M.~Pigliucci, ``Is evolvability evolvable?'' \emph{Nature Reviews Genetics},
  vol.~9, no.~1, pp. 75--82, 2008.

\bibitem{altenberg1994evolution}
L.~Altenberg \emph{et~al.}, ``The evolution of evolvability in genetic
  programming,'' \emph{Advances in genetic programming}, vol.~3, pp. 47--74,
  1994.

\bibitem{mouret2010sferes}
J.-B. Mouret and S.~Doncieux, ``Sferes v2: Evolvin'in the multi-core world,''
  in \emph{Evolutionary Computation (CEC), 2010 IEEE Congress on}.\hskip 1em
  plus 0.5em minus 0.4em\relax IEEE, 2010, pp. 1--8.

\bibitem{shaffer1995multiple}
J.~P. Shaffer, ``Multiple hypothesis testing,'' \emph{Annual review of
  psychology}, vol.~46, no.~1, pp. 561--584, 1995.

\bibitem{alexandrescu2001modern}
A.~Alexandrescu, \emph{Modern C++ design: generic programming and design
  patterns applied}.\hskip 1em plus 0.5em minus 0.4em\relax Addison-Wesley,
  2001.

\end{thebibliography}
%

\end{document}